\ifdefined\pdfminorversion\pdfminorversion=7\fi 
\ifdefined\pdfsuppresswarningpagegroup\pdfsuppresswarningpagegroup=1\fi 
\documentclass{article}

 \usepackage[main, final]{neurips_2026}

\usepackage[utf8]{inputenc} 
\usepackage[T1]{fontenc}    
\usepackage{hyperref}       
\usepackage{url}            
\usepackage{booktabs}       
\usepackage{amsfonts}       
\usepackage{nicefrac}       
\usepackage{silence}        
\usepackage{microtype}      
\usepackage{xcolor}         

\usepackage{mathtools}
\usepackage{comment}
\usepackage{graphicx}
\usepackage{color} 
\usepackage{pifont} 
\usepackage{xspace} 
\usepackage{wrapfig,lipsum} 
\usepackage{colortbl} 
\usepackage{multirow}
\usepackage{amsthm} 
\usepackage{amsmath} 
\usepackage{amssymb}
\usepackage{bm}
\usepackage{float} 
\usepackage{comment}
\usepackage{adjustbox}
\usepackage{wrapfig,lipsum,booktabs}
\usepackage{caption}
\usepackage[most]{tcolorbox}
\usepackage{xr}
\usepackage{subcaption} 
\usepackage{algorithm}
\usepackage{algpseudocode}
\usepackage{enumitem}
\usepackage{makecell}

\newtheorem{proposition}{Proposition}[section]

\newcommand{\cmarkg}{\textcolor{green}{\ding{51}}\xspace}%
\newcommand{\xmarkg}{\textcolor{red}{\ding{55}}\xspace}%

\theoremstyle{definition}

\newtheoremstyle{nonitalicremark}
  {\topsep}
  {\topsep}
  {\normalfont}
  {}
  {\bfseries}
  {.}
  { }
  {}

\theoremstyle{nonitalicremark}
\newtheorem{remark}{Remark}
\theoremstyle{plain}

\theoremstyle{plain}

\newtheorem{assumption}{Assumption}[section]

\newcommand\blfootnote[1]{%
  \begingroup
  \renewcommand\thefootnote{}%
  \NoHyper\footnote{#1}\endNoHyper%
  \addtocounter{footnote}{-1}%
  \endgroup
}

\title{FLINT: Coupling Proximal Initialization and Bounded Stochasticity for Flow-Matching Inverse Problems}

\author{%
Junseo Bang$^{1}$ \quad Daewon Choi$^{2}$ \quad \textbf{Dong Ju Mun}$^1$ \quad \textbf{Se Young Chun}$^{1,2,3\dag}$ \\
$^1$Dept. of ECE \; $^2$IPAI \; $^3$INMC \& AIIS, Seoul National University, Republic of Korea\\
\texttt{\{qkdwnstj10, eodnjschl, dongju1126, sychun\}@snu.ac.kr}
}

\begin{document}

\maketitle

\begin{abstract}
Flow matching models have emerged as a dominant paradigm for generative posterior sampling in imaging inverse problems. Existing solvers follow two distinct lines of work: trajectory-space methods, which refine sampling dynamics from random initial noise, and source-space methods, which optimize the initial noise itself. While most prior work treats these as independent design choices, we show that they are inherently coupled. We derive a unified stability bound that links initialization proximity and cumulative sampling stochasticity through a multiplicative envelope, complemented by two empirical observations. First, we identify a sufficiency regime in which a moderately optimized proximal source initialization is enough to attain the performance plateau, rendering exhaustive source-space optimization unnecessary. Second, this initial advantage is fragile: it is eroded by excessive stochasticity accumulation during guided sampling, necessitating a mechanism that bounds the cumulative noise injection. These findings motivate FLINT, a two-stage framework that operationalizes this coupling. FLINT establishes a high-fidelity starting point via a manifold-constrained proximal objective in only a few iterations, then performs guided sampling under a bounded stochasticity schedule that helps prevent erasure of the initialization advantage. Across diverse inverse problems, FLINT delivers state of the art perceptual quality and competitive distortion metrics while maintaining superior computational efficiency.
\end{abstract}

\section{Introduction}
\label{sec:intro}
\blfootnote{$\dagger$ corresponding author.}
Inverse problems arise throughout computational imaging and computer vision~\cite{bertero2021introduction, kaipio2005statistical, ongie2020deep, mccann2017convolutional, zhang2017beyond}, where the goal is to recover a clean image $\mathbf{x}_0 \in \mathbb{R}^d$ from a degraded measurement $\bm{y} = \mathcal{A}\bm{x}_0 + \mathbf{n}$, with $\mathcal{A}$ denoting the forward operator and $\mathbf{n} \sim \mathcal{N}(0, \sigma_y^2 \mathbf{I})$ additive Gaussian noise. Since $\mathcal{A}$ is typically many-to-one, the problem is ill-posed and requires a prior over natural images~\cite{tikhonov1963solution,rudin1992nonlinear,venkatakrishnan2013plug,ulyanov2018deep}. From a Bayesian perspective, this reduces to sampling from the posterior $p(\mathbf{x}_0 \mid \mathbf{y}) \propto p(\mathbf{y} \mid \mathbf{x}_0) \cdot p(\mathbf{x}_0)$, where the likelihood $p(\mathbf{y} \mid \mathbf{x}_0)$ encodes the forward model and the prior $p(\mathbf{x}_0)$ regularizes solutions toward the natural image manifold. Deep generative models have emerged as powerful priors with diffusion models~\cite{ho2020denoising, song2020score} and flow matching models~\cite{liu2023flow, lipman2023flow}. Large-scale rectified flow models~\cite{esser2024scaling} achieve photorealistic generation quality, rendering them highly attractive for posterior sampling.

\begin{figure*}[t!]
  \centering
  \includegraphics[width=1.0\textwidth]{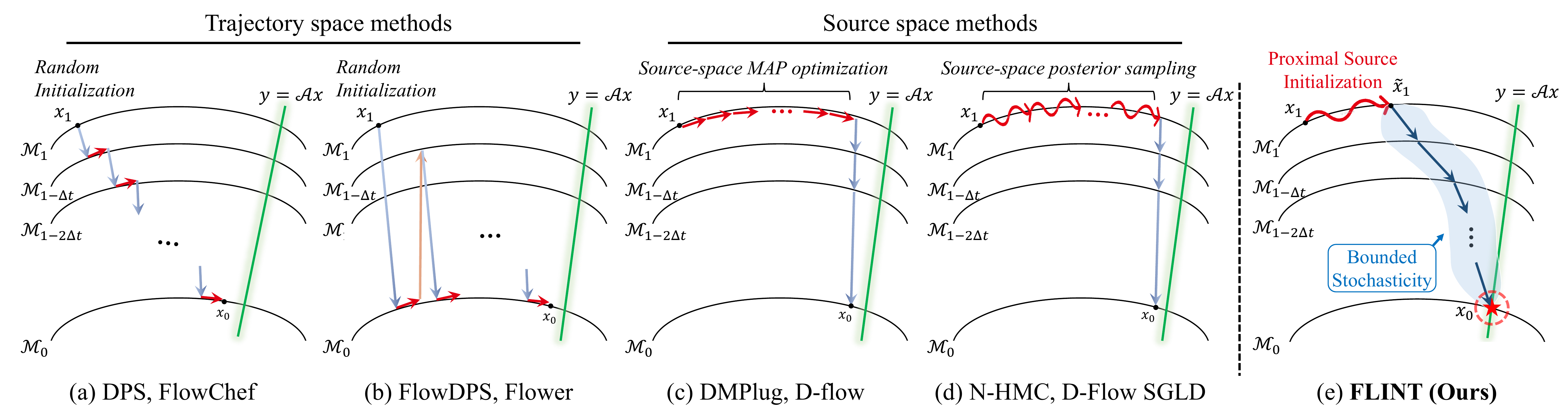}
  \vspace{-1.6em}
  \caption{
  \textbf{Illustration of generative paradigms for inverse problems}. (a) and (b) depict trajectory-space methods, while panels (c) and (d) depict source-space optimization methods. (a) Likelihood score corrections are applied directly at each noisy intermediate state $\mathbf{x}_t$. (b) Likelihood score corrections are applied in clean image space via the Tweedie estimate $\hat{\mathbf{x}}_{0|t}$, followed by renoising to the generative manifold at the next timestep. (c) Source-space MAP methods reparameterize the clean image via the deterministic generative map $\Phi$ as $\mathbf{x}_0 = \Phi(\mathbf{x}_1)$ and optimize over the initial noise $\mathbf{x}_1$ to minimize data fidelity. (d) Source-space posterior sampling draws samples from $p(\mathbf{x}_1 \mid \mathbf{y})$ via Hamiltonian Monte Carlo or Langevin dynamics. (e) By strategically coupling proximal source initialization with bounded stochasticity during guided posterior sampling, FLINT bounds the sampling trajectory, preserving the initialization advantage and yielding high-fidelity convergence.
  }
  \label{fig:introduction}
\end{figure*}

Existing inverse problem solvers leveraging these priors typically fall into two categories (Sec.~\ref{subsec:preliminary_2}). \emph{Trajectory-space methods}~\cite{dps, pigdm, ddrm, resample, daps, kim2025flowdps, flair, flower, kim2026fastdips, flowlps, trips} steer the sampling path toward the measurement-consistent manifold at each step but rely on a fixed, random starting point $\bm{x}_1 \sim \mathcal{N}(\mathbf{0}, \mathbf{I})$, leaving reconstruction quality inherently sensitive to the specific realization of the initial noise (Fig.~\ref{fig:introduction}(a) and (b)). \emph{Source-space methods}~\cite{chihaoui2024blind, wang2024dmplug, benhamu2024dflow, purohit2024posterior, kalaivanan2025ess, wang2026sourceguided, xia2026noiseadaptive, parikh2026d} perform inference directly on the initial noise $\bm{x}_1$, keeping the subsequent generative trajectory unmodified (Fig.~\ref{fig:introduction}(c) and (d)). 
Most prior work has treated initialization and sampling dynamics as decoupled modules, an implicit assumption that overlooks the coupling between initialization proximity and sampling stochasticity, leaving their joint impact on reconstruction quality largely unexplored.

In this work, we reveal that these two design choices are not independent but inherently coupled, and that exploiting their synergy is beneficial for high-fidelity posterior sampling. 
Combining a stability bound that decomposes into a source-space initialization term and a cumulative stochasticity term with empirical observations, we identify two complementary findings. 
First, a proximal source initialization that anchors the starting noise within a sufficient proximity of a source-space solution can attain high-quality reconstruction without exhaustive optimization.
Second, the benefits of this proximal start are fragile. They are eroded by the excessive accumulation of stochasticity during guided posterior sampling, necessitating a mechanism to bound the cumulative injected noise. Together, these findings lead to \emph{FLINT} (\underline{Fl}ow-based posterior sampling with proximal source \underline{in}itialization and bounded s\underline{t}ochasticity), a two-stage framework that operationalizes this coupling. Stage 1 implements a well-informed starting point by optimizing a manifold-constrained proximal objective for only a few iterations. Stage 2 then employs a bounded stochasticity schedule to steer the trajectory while preventing the erasure of the informed initialization (Fig.~\ref{fig:introduction}(e)). Our main contributions are:
\begin{itemize}[leftmargin=*, labelsep=0.5em, itemsep=0.1em, topsep=0.5em]
    \vspace{-0.5em}
    \item 
    \textbf{Theoretical Coupling:} We derive a unified stability bound for flow-based posterior sampling, showing that the terminal deviation from a deterministic reference is controlled by initialization proximity and cumulative sampling stochasticity. We empirically identify a sufficiency regime, where moderate source-space optimization is enough to reach the performance plateau.
    \item \textbf{FLINT Framework:} We introduce FLINT, a training-free two-stage sampler that bridges source-space and trajectory-space methods. It utilizes a manifold-constrained proximal objective for informed initialization and a bounded stochasticity schedule that helps preserve the initialization advantage during guided posterior sampling.
    \item 
    \textbf{Experimental Results:} Extensive experiments across imaging inverse problems show that FLINT achieves state-of-the-art perceptual quality and competitive or superior distortion metrics, with a markedly favorable efficiency-performance profile.
    \vspace{-0.5em}
\end{itemize}

\section{Background} \label{sec:preliminary}
\subsection{Rectified Flow} \label{subsec:preliminary_1}
Rectified Flow~\cite{liu2023flow} transports between the data distribution $\pi_0$ and the noise distribution $\pi_1 = \mathcal{N}(\mathbf{0}, \mathbf{I})$ via an ODE, where $t=1$ corresponds to pure noise and $t=0$ to the data space. Training pairs $(\bm{x}_0, \bm{x}_1) \sim \pi_0 \times \pi_1$ are linearly interpolated as $\bm{x}_t = (1-t)\bm{x}_0 + t\bm{x}_1$, and a velocity network $\bm{v}_\theta$ is trained to match the straight-line direction $(\bm{x}_1 - \bm{x}_0)$ as follows.
\begin{equation}
\min_\theta \mathbb{E}_{t \sim \mathcal{U}[0, 1],\, \bm{x}_0 \sim \pi_0,\, \bm{x}_1 \sim \pi_1} 
\left[ \left\| (\bm{x}_1 - \bm{x}_0) - \bm{v}_\theta(\bm{x}_t, t) \right\|_2^2 \right].
\end{equation}
Samples are generated by integrating the ODE from $t=1$ to $t=0$ over a fixed schedule $\{\tau_i^{(N)}\}_{i=1}^{N}$ with $N \in [28, 50]$ steps in practice. We denote the resulting deterministic generative map as $\Phi: \mathbb{R}^d \to \mathbb{R}^d$, so that $\bm{x}_0 = \Phi(\bm{x}_1)$. Rearranging the linear interpolation at any intermediate state $\bm{x}_t$ yields a flow-based Tweedie's formula $\hat{\bm{x}}_{0|t} := \mathbb{E}[\bm{x}_0 \mid \bm{x}_t] \approx \bm{x}_t - t\,\bm{v}_\theta(\bm{x}_t, t)$, which provides a single step clean image estimate from any intermediate state~\cite{flowdps}.

\subsection{Posterior Sampling for Inverse Problems} \label{subsec:preliminary_2}
Given the measurement model $\bm{y} = \mathcal{A}\bm{x}_0 + \mathbf{n}$, posterior sampling aims to draw samples from $p(\bm{x}_0 \mid \bm{y}) \propto p(\bm{y} \mid \bm{x}_0)\,p(\bm{x}_0)$~\cite{song2020score, song2022solving, dps, pigdm, chung2022improving}. Existing solvers built on diffusion or flow matching priors divide into two families. \emph{Trajectory-space methods} steer intermediate states toward measurement consistency via the Tweedie estimate $\hat{\bm{x}}_{0|t} := \mathbb{E}[\bm{x}_0 \mid \bm{x}_t]$, obtaining the posterior mean either via Tweedie's formula or by integrating a probability flow ODE from $\bm{x}_t$ toward $t=0$. Within this family, one subfamily enforces data consistency directly in the noisy latent space by approximating $\nabla_{\bm{x}_t}\log p(\bm{y} \mid \bm{x}_t) \approx \nabla_{\bm{x}_t}\log p(\bm{y} \mid \hat{\bm{x}}_{0|t})$ ~\cite{dps, pigdm, ddrm, blinddps, moment_matching, psld, flowchef, dilack}, while the other enforces consistency in clean image space via $\hat{\bm{x}}_{0|t}$ and then renoises back to the $(t\!+\!\Delta t)$-manifold of the generative prior~\cite{resample, daps, treg, kim2025flowdps, flair, flower, flowlps, kim2026fastdips, trips}. Both subfamilies initialize from $\bm{x}_1 \sim \mathcal{N}(\mathbf{0}, \mathbf{I})$, leaving the quality of the starting point entirely uncontrolled. \emph{Source-space methods} instead confine to the initial noise $\bm{x}_1$ via the deterministic generative map $\Phi: \mathbb{R}^d \to \mathbb{R}^d$ defined by the full reverse ODE, leaving the generative trajectory unmodified. One line of work casts source-space inference as MAP estimation~\cite{chihaoui2024blind, wang2024dmplug, benhamu2024dflow}, reparameterizing $\bm{x}_0 = \Phi(\bm{x}_1)$ and optimizing $\bm{x}_1$ to minimize data fidelity. Another targets $p(\bm{x}_1 \mid \bm{y})$ directly via Langevin dynamics, Hamiltonian Monte Carlo, or related MCMC updates~\cite{purohit2024posterior, kalaivanan2025ess, wang2026sourceguided, xia2026noiseadaptive, parikh2026d}, enabling broader posterior exploration. Both subfamilies nonetheless share a substantial computational burden: evaluating $\nabla_{\bm{x}_1} \mathcal{J}(\Phi(\bm{x}_1))$ requires backpropagating through the full $N$-step reverse ODE.
 
\subsection{Reducing the Computational Cost of Source-Space Optimization} \label{subsec:preliminary_3}
Source-space methods optimize $\bm{x}_1$ by minimizing a data consistency objective $\mathcal{J}(\bm{x}_0)$. Computing $\nabla_{\bm{x}_1} \mathcal{J}(\bm{x}_0)$ incurs two compounding costs: obtaining $\bm{x}_0$ requires an $N$-step reverse ODE, and backpropagating through this chain requires evaluating the Jacobian product $\prod_{i} \frac{\partial \bm{x}_{i-1}}{\partial \bm{x}_{i}}$.

\paragraph{Truncated sampling.} \label{par:preliminary_3_1}
To reduce the forward pass cost, we adopt truncated sampling with $K \ll N$ steps~\cite{shehata2025improved,lee2025truncated}. The flow ODE is integrated for the first $K\!-\!1$ steps along the $N$-step time schedule $\{\tau_i^{(N)}\}_{i=1}^{N}$, after which a clean image estimate is formed via the flow-based Tweedie's formula:
\begin{equation}
    \label{eq:truncated_tweedie}
    \bm{x}_{t_{\text{stop}}} = \bm{x}_1 - \textstyle\sum_{i=1}^{K-1} (\tau_i - \tau_{i+1})\,\bm{v}_\theta(\bm{x}_{\tau_i},\, \tau_i) \quad
    \hat{\bm{x}}_0 = \bm{x}_{t_{\text{stop}}} - t_{\text{stop}} \cdot \bm{v}_\theta(\bm{x}_{t_{\text{stop}}},\, t_{\text{stop}}),
\end{equation}
where $\tau_i=\tau_i^{(N)}$ and $t_{\text{stop}} = \tau_{K-1}$. Denoising to $t_{\text{stop}}$ before applying Tweedie's formula yields an efficient surrogate $\hat{\bm{x}}_0(\bm{x}_1)$.

\paragraph{Efficient gradient computation.} \label{par:preliminary_3_2}
The backpropagation cost has been mitigated by strategies such as adjoint methods and gradient checkpointing~\cite{pan2024adjointdpm, wallace2023end, chen2016training, prabhudesai2023aligning, clark2024directly}, though these often trade memory for computation. Shortcut Diffusion Optimization (SDO)~\cite{dou2025sdo} reinterprets sampling as Picard fixed-point iteration and shows that $\nabla_{\mathbf{x}_1} \mathcal{J}(\mathbf{x}_0)$ is well-approximated by $\frac{\partial \mathcal{J}}{\partial \mathbf{x}_0} \frac{\partial \mathbf{x}_{N-1}}{\partial \mathbf{x}_1}$, retaining the computational graph only at the first denoising step with bounded error. Our method combines truncated sampling and the SDO shortcut, as detailed in Algorithm~\ref{alg:sdo_euler}, to substantially reduce per-iteration overhead.

\begin{figure*}[t!]
  \centering
  \includegraphics[width=0.78\textwidth]{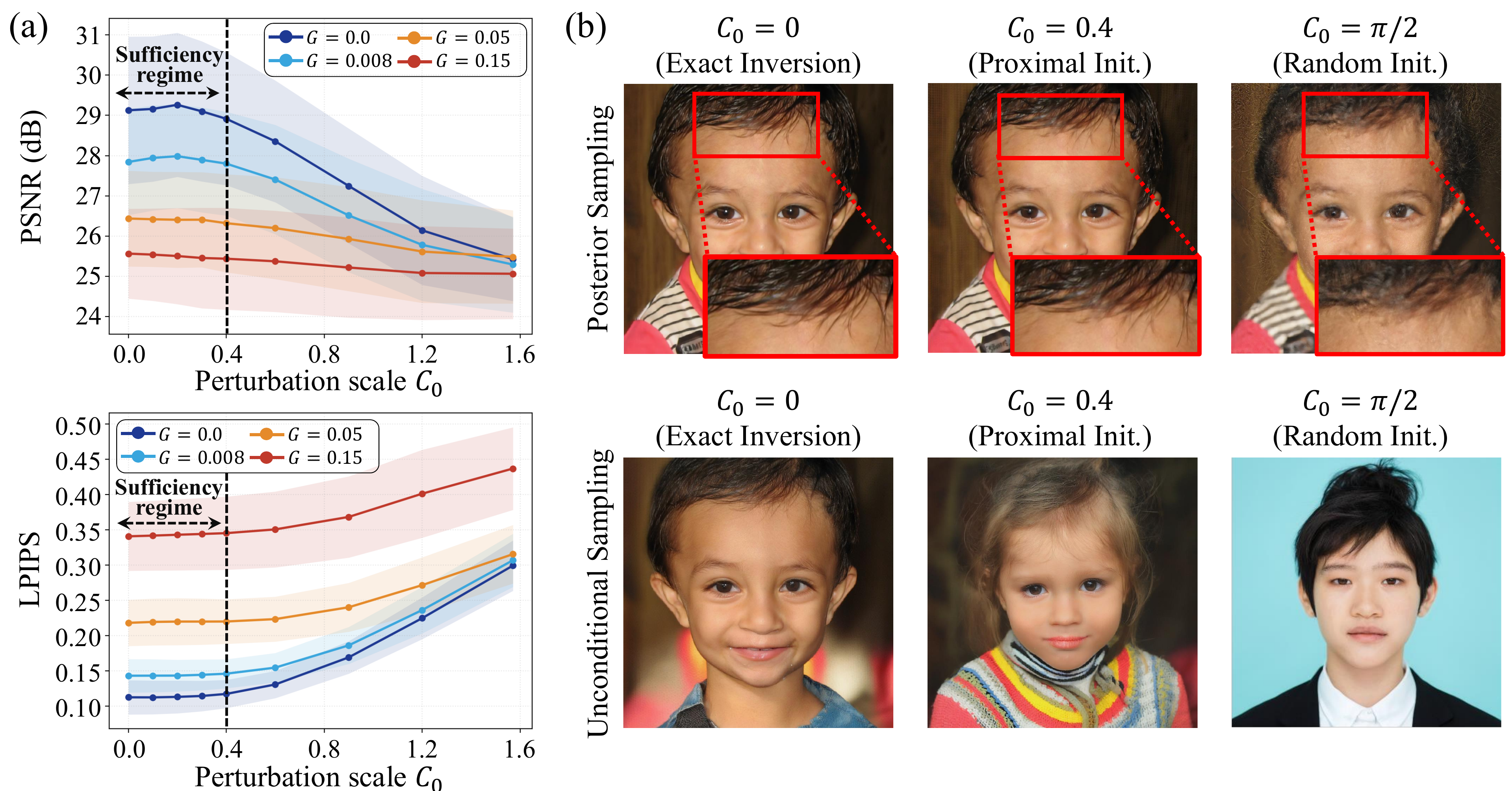}
  \vspace{-0.5em}
  \caption{
  \textbf{Interplay of proximal initialization and sampling stochasticity} on Gaussian deblurring.
  (a) PSNR and LPIPS as functions of the perturbation scale $C_0$ across varying the cumulative stochasticity $G$, demonstrating the existence of a sufficiency regime ($C_0 \le 0.4$ in this setup) where proximal initialization performs on par with exact inversion.
  (b) Visual comparisons for $C_0 \in \{0, 0.4, \pi/2\}$. While a representative proximal initialization within this regime ($C_0 = 0.4$) matches the high fidelity of exact inversion under posterior sampling ($G=0.008$), it preserves measurement-consistent structures even in unconditional sampling, unlike the unrelated output from random initialization.
  }
  \label{fig:analysis_1}
\end{figure*}

\section{Analysis: Coupling of Proximal Initialization and Sampling Stochasticity}
\label{sec:analysis}
This section analyzes how source-space proximity and sampling stochasticity jointly affect an upper bound on the mean-square terminal deviation from a reference trajectory during guided sampling. 
An initial noise placed near the minimizer of the posterior energy $\mathcal{E}(\bm{x}_1) \coloneqq -\log p(\bm{x}_1 \mid \bm{y})$ admits an energy upper bound controlled by its distance to the minimizer, whereas a random Gaussian draw typically lies far from this minimizer in high dimensions and thus yields a looser distance-based upper bound.
This argument is formalized in Appendix~\ref{app:subsec_energy_advantage}. 
Whether such an initialization advantage is retained during sampling depends on how the proximity gap propagates through the sampling trajectory, which we address by analyzing the reverse-time posterior flow SDE~\cite{anderson1982reverse,song2020score}:
\begin{equation}
  \label{eq:governing_sde}
  d\bm{x}_t
  \;=\;
  \left[
    v_t(\bm{x}_t \mid \bm{y})
    - \frac{g_t^2}{2}
      \nabla_{\bm{x}_t}\!\log p_t(\bm{x}_t \mid \bm{y})
  \right]dt
  \;+\;
  g_t\,d\bar{\bm{W}}_t,
\end{equation}
where $v_t(\bm{x}_t \mid \bm{y})$ is the posterior velocity, $\nabla_{\bm{x}_t}\!\log p_t(\bm{x}_t \mid \bm{y})$ is the posterior score, and $g_t$ is the diffusion coefficient controlling stochastic departure.

\begin{proposition}[Terminal stability bound]
\label{prop:error_decomposition}
Let $\bm{x}_1^* \in \mathbb{R}^d$ be any reference point in the $d$-dimensional source space, $\bm{x}_0^*$ the terminal output of the drift ODE of~\eqref{eq:governing_sde} with only the Brownian term removed which is initialized at $\bm{x}_1^*$, and $\Phi_g(\bm{x}_1)$ the terminal output of~\eqref{eq:governing_sde} initialized at $\bm{x}_1$. Let $G \coloneqq \int_0^1 g_t^2\,dt$ denote the cumulative stochasticity. Suppose the posterior velocity satisfies a one-sided Lipschitz condition with parameter $m_v(t)$ (Assumption~\ref{assum:onesided_vel}) and the posterior score satisfies a one-sided Lipschitz condition with parameter $m_s(t)$ (Assumption~\ref{assum:onesided_score}). Then any $\tilde{\bm{x}}_1 \in \mathbb{R}^d$ satisfying $\|\tilde{\bm{x}}_1 - \bm{x}_1^*\|_2 \leq \delta$ obeys:
\begin{equation}
  \mathbb{E}\!\left[
    \|\Phi_g(\tilde{\bm{x}}_1) - \bm{x}_0^*\|_2^2
  \right]
  \leq
  2\,e^{2 M_+}
  \!\left(
    \underbrace{\delta^2}_{\smash{\substack{\text{initialization} \\ \text{error term}}}}
    \,+\,
    \underbrace{d \cdot G}_{\smash{\substack{\text{stochastic} \\ \text{departure term}}}}
  \right),
  \;
  M_+ \coloneqq \max\!\Bigl(0,\;\sup_{s \in [0,1]} \int_0^s m(u)\,du\Bigr),
  \label{eq:error_bound}
\end{equation}
where $m(t) \coloneqq m_v(t) + \tfrac{g_t^2}{2}\,m_s(t)$ is the effective contraction rate of the posterior drift. \textup{(Proof in Appendix~\ref{app:subsec_proof_error_decomposition}.)}
\end{proposition}

\begin{remark}[Canonical reference: prior-typical and measurement-consistent initialization]
\label{remark:principled_init}
A canonical instantiation of Proposition~\ref{prop:error_decomposition} selects $\bm{x}_1^*$ as a reference point that is simultaneously prior typical and measurement consistent, satisfying $A(\Phi(\bm{x}_1^*)) \approx \bm{y}$ under the unconditional generative map $\Phi$. Proposition~\ref{prop:error_decomposition} neither bounds reconstruction error relative to the ground truth nor establishes convergence to the exact posterior.
\end{remark}

\paragraph{Theoretical insights.}
Proposition 3.1 places the initialization error and the stochastic departure in a single additive bound under a common prefactor $e^{2M_+}$, yielding two key insights independent of the choice of reference.
\emph{(i) Sufficiency of proximal initialization.} When $\delta^2 \ll d \cdot G$, the stochastic departure dominates the bound, so further reducing $\delta$ has only a small relative effect on the upper bound. Achieving sufficient proximity thus controls the initialization term adequately without exhaustive optimization. 
\emph{(ii) Attenuation by excessive stochasticity.} Increasing $G$ inflates the stochastic departure term $d \cdot G$, so that, for a fixed $M_+$, the bound grows with $G$ regardless of how well $\delta$ has been reduced. Bounding $G$ throughout sampling therefore helps preserve the benefit of any principled starting point.

We explore the two insights empirically under the setting of Remark~\ref{remark:principled_init}. Using FlowDPS~\cite{flowdps} as the posterior sampler on Gaussian deblurring (FFHQ~\cite{ffhq}, 100 samples), we instantiate $\bm{x}_1^*$ as the unconditional ODE inversion of the ground truth image and construct perturbed initializations $\tilde{\bm{x}}_1$ via spherical linear interpolation (SLERP) between $\bm{x}_1^*$ and a Gaussian noise $\bm{\epsilon} \sim \mathcal{N}(\bm{0}, \bm{I}_d)$:
\begin{equation}
  \label{eq:slerp}
  \tilde{\bm{x}}_1
  = \frac{\sin(\theta - C_0)}{\sin\theta}\,\bm{x}_1^*
  + \frac{\sin C_0}{\sin\theta}\,\bm{\epsilon},
  \qquad
  \theta = \cos^{-1}\!\left(
    \frac{\bm{x}_1^* \cdot \bm{\epsilon}}
         {\|\bm{x}_1^*\|_2 \|\bm{\epsilon}\|_2}
  \right),
\end{equation}
where $C_0 \in [0, \pi/2]$ controls the perturbation scale. 
This construction approximately preserves the Gaussian typical norm $\|\tilde{\bm{x}}_1\|_2\approx\sqrt d$, providing a clean empirical counterpart to the initialization error term in~\eqref{eq:error_bound}. Figure~\ref{fig:analysis_1}(a) visualizes the perception-distortion landscape as a joint function of the perturbation scale $C_0$, reflecting $\delta^2$ along the horizontal axis, and the cumulative stochasticity $G$ across the curve family.

\paragraph{Sufficiency of proximal source initialization.} \label{par:analysis_1}
Across all $G$ curves in Figure~\ref{fig:analysis_1}(a), reconstruction quality remains nearly flat within an initial proximity range before it begins to degrade. The exact extent of this flat region (here, up to $C_0 \approx 0.4$) depends on the specific setup. We refer to this empirically observed region as the ``sufficiency regime''.
Within this regime, further reducing $\delta$ provides negligible gain. Figure~\ref{fig:analysis_1}(b) qualitatively corroborates this. Under posterior sampling ($G = 0.008$), a moderately proximal initialization from this regime (e.g., $C_0 = 0.4$) achieves reconstructions visually comparable to exact inversion ($C_0 = 0$), whereas random initialization ($C_0 = \pi/2$) incurs a substantial quality drop. Unconditional sampling further confirms that the proximal noise encodes measurement-consistent structure, while the random noise produces an unrelated image.

\paragraph{Attenuation by excessive stochasticity.} \label{par:analysis_2}
Turning to the curve family in Figure~\ref{fig:analysis_1}(a), increasing $G$ degrades reconstruction quality. This is qualitatively consistent with Proposition~\ref{prop:error_decomposition}: as $G$ grows, the stochastic departure term $d \cdot G$ rises directly, so the bound increases regardless of how well $\delta$ has been reduced. Consequently, a well chosen initialization alone is insufficient. Bounding $G$ throughout guided sampling helps preserve the advantage conferred by proximal noise initialization. Additional empirical validation is provided in Appendix~\ref{app:more_analysis}.

\section{Method: The FLINT Framework} \label{sec:method}
The analysis in Sec.~\ref{sec:analysis} motivates the two-stage design of FLINT. Since exhaustive source space optimization is redundant under the sufficiency regime (Sec.~\ref{par:analysis_1}), Stage~1 (Sec.~\ref{subsec:method_1}) efficiently anchors the initial noise via a manifold constrained proximal objective with spectrally weighted data consistency. Stage~2 (Sec.~\ref{subsec:method_2}) then employs a bounded stochasticity schedule during guided posterior sampling, preventing excessive stochasticity from eroding the fidelity attained by this initialization (Sec.~\ref{par:analysis_2}).

\subsection{Stage 1: Proximal Source Initialization} \label{subsec:method_1}
\paragraph{Manifold-constrained proximal objective.} \label{par:method_1_1}
Unlike source space MAP methods that incur massive computational costs for a deterministic exact solution, or MCMC approaches that exhaustively characterize the source posterior distribution, Stage~1 focuses solely on rapidly attaining sufficient proximity, a measurement-informed initialization statistically consistent with the generative prior. To derive a principled objective, we employ a variational relaxation, minimizing the KL divergence from a surrogate distribution $q_\phi(\bm{x}_1)=\mathcal{N}(\bm{x}_1\mid \bm{\mu}_\phi,\rho^2\mathbf{I})$ to the intractable source posterior $p(\bm{x}_1\mid\bm{y})$:
\begin{equation}
\label{eq:source_kl}
\begin{gathered}
  \min_{\phi}\;
    \mathrm{KL}\!\left(q_\phi(\bm{x}_1)\;\|\;p(\bm{x}_1\mid\bm{y})\right)\\
  = \underbrace{-\mathbb{E}_{q_\phi}\!\left[\log p(\bm{y}\mid\bm{x}_1)\right]}_{\text{data fidelity}}
     + \underbrace{\mathrm{KL}\!\left(q_\phi(\bm{x}_1)\;\|\;p(\bm{x}_1)\right)}_{\text{prior regularization}}
     + \log p(\bm{y})
\end{gathered}
\end{equation}
where $\log p(\bm{y})$ is constant. Crucially, the standard KL term only constrains the global norm of $\bm{\mu}_\phi$ but cannot preclude higher-order departures from Gaussianity in source space. To encourage the optimized noise adheres to the prior manifold, we augment the objective with a higher order cumulant regularizer $\mathcal{R}(\bm{\mu}_\phi)$ that penalizes non-Gaussian drift during optimization (Appendix~\ref{app:details_flint_1_3}).

\begin{algorithm}[t]
\caption{The FLINT solver for flow-based inverse problems}
\label{alg:flint}
\footnotesize
\begin{minipage}[t]{0.48\linewidth}
\textbf{Stage 1: Source-Space Proximal Initialization}
\begin{algorithmic}[1]
\Require $\bm{y}$, $\mathcal{A}$, $v_\theta$, $\beta_{\mathrm{KL}}$, $\lambda_{\mathrm{cum}}$
\State Initialize $\bm{\mu}_\phi \sim \mathcal{N}(\bm{0}, \mathbf{I})$
\For{$s = 1, \ldots, S$}
    \State $\bm{x}_1 \leftarrow \bm{\mu}_\phi + \rho^{(s)} \bm{\epsilon}$,\;
           $\bm{\epsilon} \sim \mathcal{N}(\bm{0}, \mathbf{I})$
    \State $\hat{\bm{x}}_0 \leftarrow \widehat{\Phi}_{\mathrm{tr}}(\bm{x}_1)$
           \hfill\Comment{Algorithm~\ref{alg:sdo_euler}}
    \State $\mathcal{J} \leftarrow \ell_{\mathrm{IR}}(\bm{y}, \hat{\bm{x}}_0)
           + \beta_{\mathrm{KL}}\,\mathrm{KL}(q_\phi(\bm{x}_1) \| p(\bm{x}_1))
           + \lambda_{\mathrm{cum}}\,\mathcal{R}(\bm{\mu}_\phi)$
           \hfill\Comment{$\ell_{\mathrm{IR}}$: Eq.~\eqref{eq:spectral_dc}}
    \State $\bm{\mu}_\phi \leftarrow \mathrm{Adam}(\nabla_{\bm{\mu}_\phi}\mathcal{J})$
\EndFor
\State \Return $\tilde{\bm{x}}_1 = \bm{\mu}_\phi$
\end{algorithmic}
\end{minipage}
\hfill\vline\hfill
\begin{minipage}[t]{0.48\linewidth}
\textbf{Stage 2: Proximity-Preserving Posterior Sampling with Bounded Stochasticity}
\begin{algorithmic}[1]
\Require $\tilde{\bm{x}}_1$, $\bm{y}$, $\mathcal{A}$, $v_\theta$, $\lambda$, $c$, $c_\emptyset$, $\eta(t)$ \Comment{$\eta(t)$: Eq.~\eqref{eq:eta_schedule}}
\State $\bm{x}_1 \leftarrow \tilde{\bm{x}}_1$
\For{$t : 1 \to 0$}
    \State $v_t \leftarrow v_\theta(\bm{x}_t, c_\emptyset)
           + \lambda\,\bigl(v_\theta(\bm{x}_t, c)
           - v_\theta(\bm{x}_t, c_\emptyset)\bigr)$
    \State $\hat{\bm{x}}_{0|t} \leftarrow \bm{x}_t - \sigma_t v_t$,\;
           $\hat{\bm{x}}_{1|t} \leftarrow \bm{x}_t + (1-\sigma_t) v_t$
    \State $\tilde{\bm{x}}_{0|t} \leftarrow
       \underset{\bm{x}}{\arg\min}\;
       \mathcal{L}\!\bigl(\mathcal{A}(\bm{x}),\,\bm{y}\bigr)$
       \hfill\Comment{init.\ at $\hat{\bm{x}}_{0|t}$}
    \State $\bm{\epsilon} \sim \mathcal{N}(\bm{0}, \mathbf{I})$
    \State $\bm{x}_{t+\Delta t} \leftarrow
           (1-\sigma_{t+\Delta t})\tilde{\bm{x}}_{0|t}$
           $+ \,\sigma_{t+\Delta t}\hat{\bm{x}}_{1|t}
           + \eta(t)\bm{\epsilon}$
\EndFor
\State \Return $\bm{x}_0$
\end{algorithmic}
\end{minipage}
\end{algorithm}

Combining this regularizer with the computationally efficient truncated sampling $\widehat{\Phi}_{\mathrm{tr}}$ (Sec.~\ref{par:preliminary_3_2}) and the data consistency loss (Sec.~\ref{par:method_1_2}), we formulate the manifold-constrained proximal objective:
\begin{equation}
\label{eq:source_proximal}
    \mathcal{J}(\phi) = \mathbb{E}_{\bm{x}_1 \sim q_\phi}\!\left[
        \ell_{\mathrm{IR}}(\bm{y},\, \widehat{\Phi}_{\mathrm{tr}}(\bm{x}_1))
    \right]
    + \beta_{\mathrm{KL}} \cdot \mathrm{KL}\!\left(
        q_\phi(\bm{x}_1) \,\|\, p(\bm{x}_1)
    \right)
    + \lambda_{\mathrm{cum}} \cdot \mathcal{R}(\bm{\mu}_\phi),
\end{equation}
where the expectation is approximated via a single sample Monte Carlo estimator $\bm{x}_1 = \bm{\mu}_\phi + \rho\bm{\epsilon}$ with a fixed annealing schedule for $\rho$. Gradients of $\mathcal{J}(\phi)$ are computed via the SDO shortcut combined with $\widehat{\Phi}_{\mathrm{tr}}$, yielding a low per-iteration cost. Within just a few iterations, this optimization yields a proximity-initialized noise, a starting point residing in a low energy posterior region.

\paragraph{Spectrally weighted data consistency.} \label{par:method_1_2}
The truncated ODE estimate $\hat{\bm{x}}_0=\widehat{\Phi}_{\mathrm{tr}}(\bm{x}_1)$ is spectrally biased: with few reverse sampling steps, high-frequency components remain largely unresolved~\cite{choi2022perception}, making data consistency gradients unreliable over the full spectrum and destabilizing the proximal optimization. 
We address this by estimating frequency dependent reliability with the Wiener filter framework~\cite{wiener1949extrapolation}. For the flow-matching interpolant $\bm{x}_{t_{\mathrm{stop}}} = (1-t_{\mathrm{stop}})\bm{x}_0 + t_{\mathrm{stop}}\boldsymbol{\epsilon}$, the signal and noise powers at frequency $\omega$ are $(1-t_{\mathrm{stop}})^2 S_{x_0}(\omega)$ and $t_{\mathrm{stop}}^2$, respectively, where $S_{x_0}(\omega) \propto \|\omega\|^{-\alpha}$ follows the well-established pink noise model for natural images~\cite{simoncelli2001natural}. Applying the Wiener filter formula directly yields the following data consistency loss:
\begin{equation}
\label{eq:spectral_dc}
\begin{aligned}
\ell_{\mathrm{IR}}(\bm{y},\hat{\bm{x}}_0)
=
\left\|
W(t_{\mathrm{stop}})\odot
\mathcal{F}
\!\left[
A^\ast
\!\left(
A\hat{\bm{x}}_0-\bm{y}
\right)
\right]
\right\|_2^2, \;
W(\omega,t_{\mathrm{stop}})
=
\frac{(1-t_{\mathrm{stop}})^2S_{x_0}(\omega)}
{(1-t_{\mathrm{stop}})^2S_{x_0}(\omega)+t_{\mathrm{stop}}^2},
\end{aligned}
\end{equation}
Here, $\mathcal{F}$ and $A^\ast$ denotes the 2D Fourier transform, and the adjoint of the forward operator, respectively. $A^\ast(A\hat{\bm{x}}_0-\bm{y})$ back-projects the measurement residual to the image domain before spectral filtering.
The weight $W \in [0,1]$ emphasizes reliable low frequencies and decays toward~0 at high frequencies. By suppressing gradients from unreliable frequency components, Eq.~\eqref{eq:spectral_dc} provides stable low-frequency supervision, yielding faster convergence than full-spectrum data consistency.


\subsection{Stage 2: Proximity-Preserving Posterior Sampling with Bounded Stochasticity} \label{subsec:method_2}
Stage~2 starts from the proximity-initialized source $\tilde{\bm{x}}_1=\bm{\mu}_\phi$ produced by Stage~1, which provides a measurement-informed starting point. Its role is not to rediscover a measurement-consistent low-frequency structure from scratch, but rather to perform the proximity-preserving stochastic refinement along the sampling trajectory. We adopt the framework of the second sub-family of trajectory-space methods described in Sec.~\ref{subsec:preliminary_2}. At each reverse step, this approach obtains a clean image estimate via the flow-based Tweedie's formula, applies a data consistency correction in the clean image space, and renoises the corrected estimate onto the generative prior manifold at the next timestep. Concretely, the next state is computed as:
\begin{equation}
    \bm{x}_{t + \Delta t}
    = (1 - \sigma_{t + \Delta t})\,\tilde{\bm{x}}_{0|t}(\bm{y})
      + \sigma_{t + \Delta t}\,\hat{\bm{x}}_{1|t}
      + \eta(t)\,\bm{\epsilon}, \quad \bm{\epsilon} \sim \mathcal{N}(\bm{0}, \mathbf{I}),
    \label{eq:stage2_update}
\end{equation}
where $\sigma_t$ denotes the noise level, and the update reinterpolates the corrected clean estimate $\tilde{\bm{x}}_{0|t}(\bm{y})$ and noise estimate $\hat{\bm{x}}_{1|t}$ back onto the $(t+\Delta t)$-manifold while injecting stochasticity scale $\eta(t)$.

\paragraph{Bounded stochasticity schedule.}
Proposition~\ref{prop:error_decomposition} motivates limiting stochastic accumulation but does not uniquely determine how the stochasticity budget should be allocated over time. We design the following stochasticity schedule:
\begin{equation}
    \eta(t) = \lambda_s \cdot \sqrt{1 - \sigma_t} \cdot \sigma_t,
    \label{eq:eta_schedule}
\end{equation}
where $\lambda_s$ controls the overall magnitude.
The shape is deliberately chosen to endow Eq.~\eqref{eq:stage2_update} with a proximity-preserving role. At early timesteps $t \approx 1$ ($\sigma_t \approx 1$), $\eta(t)$ remains close to zero, suppressing stochastic perturbations during the phase where the coarse image layout is determined in flow-based models. Large perturbations at this stage would erode the low-energy proximal state established by Stage~1. As $t$ decreases to~0, $\eta(t)$ again decays to zero, effectively concentrating stochasticity in the intermediate timesteps where perceptual details are refined without disturbing the global structure. With $\lambda_s < 1$, the bell-shaped profile of $\eta(t)$ keeps $\int_0^1 \eta^2(t)\,\mathrm{d}t$ bounded, aiming to preserve the initialization advantage throughout guided sampling. The overall algorithm is summarized in Algorithm~\ref{alg:flint}.

\begin{table*}[t!]
    \centering
    \renewcommand{\arraystretch}{1.0} 
    \setlength{\tabcolsep}{0.2em} 
    \caption{
    \textbf{Quantitative comparison on a flow matching model} averaged over 1,000 FFHQ and 800 DIV2K samples using 50 NFEs under Gaussian noise ($\sigma_y = 0.03$). 
    }
    \vspace{-0.5em}
    \resizebox{0.99\textwidth}{!}{
    \begin{tabular}{l|cccc|cccc|cccc|cccc}
    \toprule
    & \multicolumn{4}{c|}{Super-Resolution $\times$8}
    & \multicolumn{4}{c|}{Super-Resolution $\times$12}
    & \multicolumn{4}{c|}{Motion Deblurring}
    & \multicolumn{4}{c}{Gaussian Deblurring} \\
    \midrule
    Method
    & PSNR$\uparrow$ & SSIM$\uparrow$ & FID$\downarrow$ & LPIPS$\downarrow$
    & PSNR$\uparrow$ & SSIM$\uparrow$ & FID$\downarrow$ & LPIPS$\downarrow$
    & PSNR$\uparrow$ & SSIM$\uparrow$ & FID$\downarrow$ & LPIPS$\downarrow$
    & PSNR$\uparrow$ & SSIM$\uparrow$ & FID$\downarrow$ & LPIPS$\downarrow$ \\
    \midrule
    \multicolumn{17}{c}{\textbf{FFHQ (768 $\times$ 768)}}\\
    \midrule
    FlowChef & 27.94 & 0.773 & 104.19 & 0.438 &
    26.21 & 0.725 & 170.39 & 0.554 &
    25.42 & 0.733 & 88.78 & 0.478 &
    26.62 & 0.748 & 103.13 & 0.488 \\
    
    FlowDPS & 27.08 & 0.711 & 28.91 & 0.379 &
    26.66 & 0.736 & 30.55 & 0.354 &
    25.02 & 0.688 & 45.54 & 0.427 &
    25.80 & 0.699 & 47.21 & 0.420 \\
    
    FLAIR & 27.58 & 0.697 & \underline{29.73} & \underline{0.285} &
    26.77 & 0.709 & \underline{24.75} & \underline{0.307} &
    \underline{29.39} & \underline{0.755} & \textbf{13.83} & \underline{0.259} &
    \underline{28.68} & 0.744 & \underline{15.24} & \underline{0.259} \\
    
    Flower & \underline{28.36} & \underline{0.793} & 35.97 & 0.351 &
    \underline{27.15} & \underline{0.762} & 76.44 & 0.444 &
    26.38 & 0.756 & 71.93 & 0.4410 &
    26.96 & \underline{0.759} & 86.03 & 0.462 \\
    
    FlowLPS & 26.67 & 0.691 & 30.41 & 0.386 &
    25.84 & 0.685 & 62.16 & 0.429 &
    25.80 & 0.715 & 21.25 & 0.365 &
    26.49 & 0.718 & 23.46 & 0.364 \\
    
    \rowcolor{red!10}
    Ours & \textbf{29.63} & \textbf{0.795} & \textbf{21.60} & \textbf{0.217} &
    \textbf{27.74} & \textbf{0.763} & \textbf{24.26} & \textbf{0.268} &
    \textbf{30.77} & \textbf{0.799} & \underline{14.95} & \textbf{0.186} &
    \textbf{29.88} & \textbf{0.783} & \textbf{15.14} & \textbf{0.196} \\
    \midrule
    \multicolumn{17}{c}{\textbf{DIV2K (768 $\times$ 768)}}\\
    \midrule
    FlowChef & 22.58 & 0.586 & 89.97 & 0.578 &
    21.14 & \underline{0.516} & 141.43 & 0.693 &
    20.31 & 0.510 & 106.78 & 0.647 &
    21.08 & 0.517 & 133.41 & 0.675 \\
    
    FlowDPS & 21.60 & 0.511 & 39.23 & 0.503 &
    \underline{20.84} & 0.483 & 47.37 & 0.554 &
    19.79 & 0.453 & 55.84 & 0.599 &
    20.35 & 0.456 & 61.20 & 0.614 \\
    
    FLAIR & 21.11 & 0.484 & \textbf{30.70} & \underline{0.385} &
    19.92 & 0.439 & \underline{43.12} & \underline{0.461} &
    \underline{24.16} & \underline{0.644} & \underline{17.33} & \underline{0.303} &
    \underline{22.74} & \underline{0.572} & \underline{27.73} & \underline{0.378} \\
    
    Flower & \underline{22.86} & \underline{0.595} & 41.19 & 0.479 &
    \textbf{21.69} & \textbf{0.546} & 71.51 & 0.592 &
    21.16 & 0.544 & 80.38 & 0.581 &
    21.56 & 0.538 & 91.03 & 0.622 \\
    
    FlowLPS & 21.78 & 0.516 & 34.05 & 0.450 &
    20.82 & 0.479 & 58.35 & 0.507 &
    20.64 & 0.507 & 33.12 & 0.451 &
    20.98 & 0.491 & 39.14 & 0.465 \\
    
    \rowcolor{red!10}
    Ours & \textbf{23.10} & \textbf{0.598} & \underline{33.21} & \textbf{0.356} &
    20.35 & 0.496 & \textbf{35.01} & \textbf{0.432} &
    \textbf{26.55} & \textbf{0.728} & \textbf{15.69} & \textbf{0.197} &
    \textbf{24.13} & \textbf{0.629} & \textbf{25.39} & \textbf{0.301} \\
    \bottomrule
    \end{tabular}
    }
    \label{tab:main_tab_1}
\end{table*}

\section{Experiments} \label{sec:experiments}

\subsection{Experimental Setup} \label{subsec:experimental_setup}

\paragraph{Datasets and metrics.} \label{par:data_metric}
Experiments are conducted on FFHQ~\cite{ffhq} ($1024^2$, 1k samples) and DIV2K~\cite{div2k} (2K, 800 samples), with PSNR, SSIM~\cite{ssim}, LPIPS~\cite{lpips}, and patch-based FID~\cite{fid} ($256\times256$ patches, InceptionV3) as evaluation metrics. Additionally, for ablation studies conducted on a smaller subset (100 samples), we report patch-based KID~\cite{binkowski2018demystifying}.

\paragraph{Baselines.} \label{par:baselines}
We compare against recent flow matching approaches: FlowChef~\cite{flowchef}, FlowDPS~\cite{flowdps}, FLAIR~\cite{flair}, Flower~\cite{flower}, and FlowLPS~\cite{flowlps}, all built upon Stable Diffusion 3.5-Medium ($768^2$, NFE 50). For FLINT, the same 50 NFE budget is split into 21 NFEs for Stage~1 ($K{=}3$ truncated steps $\times\,7$ iterations) and 29 NFEs for Stage~2. Baseline configurations are detailed in Appendix~\ref{app:implementation}.

\paragraph{Problem setting.} \label{par:problem_setting}
We evaluate on several linear inverse problems: SR$\times$8, $\times$12, motion deblurring with a $61\times61$ kernel at intensity $0.5$, and Gaussian deblurring with a $61\times61$ kernel at standard deviation $3.0$. All tasks incorporate additive Gaussian noise with $\sigma_y=0.03$ (corresponding to $1.5\%$ of the normalized pixel range). For text conditioning, a fixed prompt ("A high quality photo of a face") is used for FFHQ, while image adaptive captions from DAPE~\cite{seesr} are used for DIV2K.

\begin{figure*}[t!]
  \centering
  \includegraphics[width=1.0\textwidth]{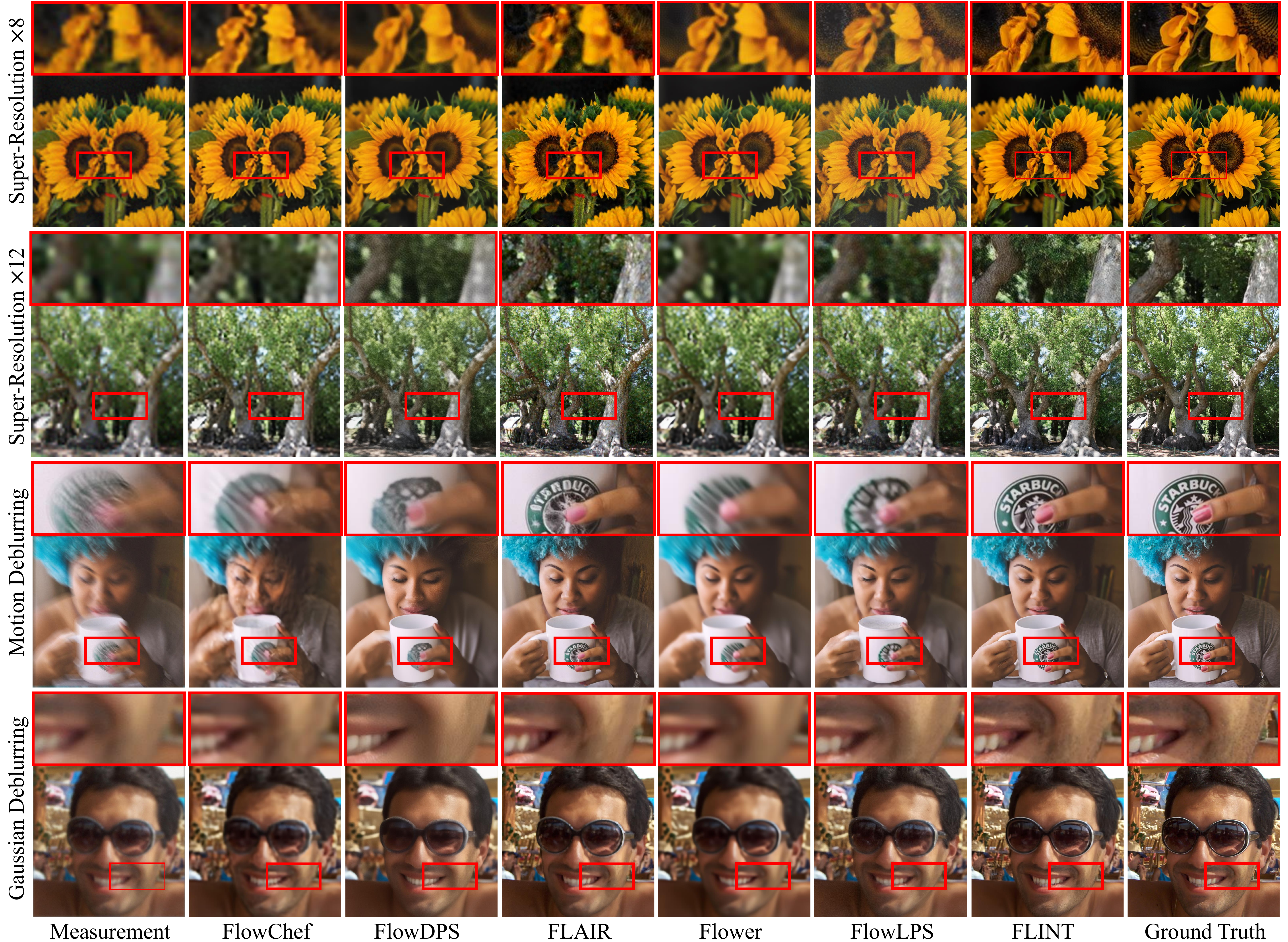}
  \caption{
  \textbf{Qualitative comparison} for FFHQ and DIV2K datasets.
  }
  \label{fig:quali_flow}
\end{figure*}

\begin{figure*}[t]
  \centering
  \includegraphics[width=0.95\textwidth]{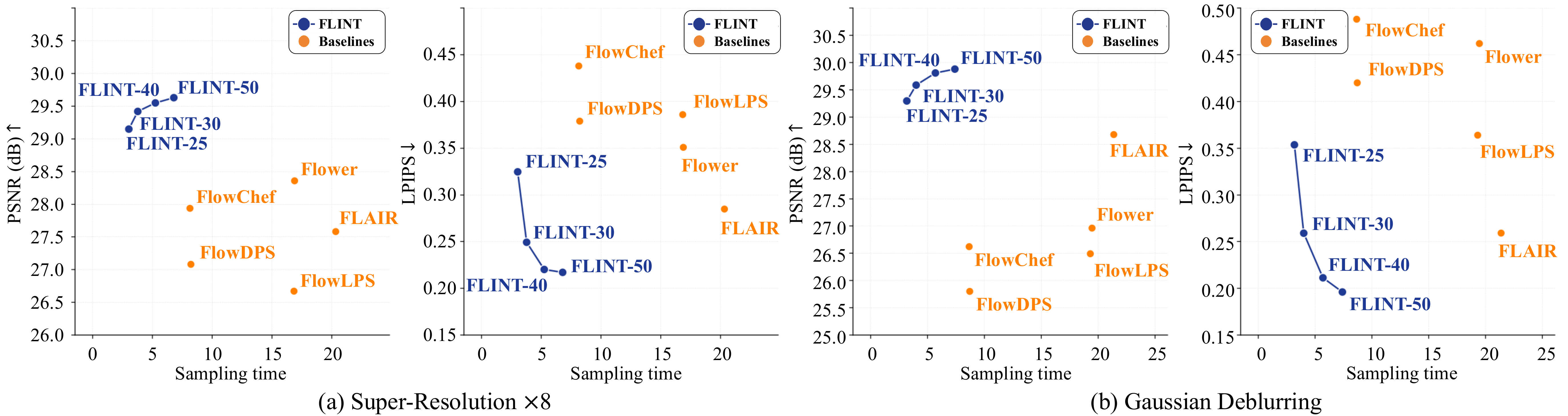}
  \caption{
        \textbf{Computational cost and performance.} PSNR and LPIPS vs. sampling time (sec) for SR$\times$8 and Gaussian deblurring. Numbers after method names denote its NFEs.
  }
  \label{fig:robustness_nfe}
\end{figure*}

\begin{table*}[t!]
    \centering
    \renewcommand{\arraystretch}{1.0} 
    \setlength{\tabcolsep}{0.6em}
    \caption{\textbf{Quantitative comparison on a diffusion model} over 1,000 FFHQ samples using 200 NFEs under additive Gaussian noise ($\sigma_y = 0.03$).}
    \vspace{-0.5em}
    \resizebox{0.8\textwidth}{!}{
        \begin{tabular}{l|cccc|cccc}
        \toprule
        & \multicolumn{4}{c|}{Super-Resolution $\times$4} & \multicolumn{4}{c}{Motion Deblurring} \\
        \midrule
        Method &
        PSNR$\uparrow$ & SSIM$\uparrow$ & FID$\downarrow$ & LPIPS$\downarrow$ &PSNR$\uparrow$ & SSIM$\uparrow$ & FID$\downarrow$ & LPIPS$\downarrow$ \\
        \midrule
        PSLD & 26.07 & 0.712 & 46.69 & 0.309 & 25.06 & 0.703 & 59.61 & 0.367 \\
        ReSample & 25.96 & \textbf{0.849} & 50.97 & \underline{0.225} & 24.18 & \underline{0.765} & 92.86 & 0.397 \\
        DDPG & \underline{29.73} & \underline{0.809} & 84.12 & 0.305 & \underline{26.62} & 0.752 & 94.41 & 0.358 \\
        TReg & 25.22 & 0.756 & \underline{37.35} & 0.244 & 25.42 & 0.701 & \underline{45.38} & \underline{0.274} \\
        \rowcolor{red!10}
        Ours & \textbf{29.88} & 0.776 & \textbf{22.36} & \textbf{0.161} & \textbf{30.05} & \textbf{0.791} & \textbf{33.32} & \textbf{0.182} \\
        \bottomrule
        \end{tabular}
    }
    \label{tab:diffusion_tab}
\end{table*}

\subsection{Experimental Results} \label{subsec:exp_results}
\paragraph{Flow matching model.} \label{par:exp_results_1}
As shown in Table~\ref{tab:main_tab_1}, FLINT consistently outperforms all flow matching baselines, yielding favorable results across both distortion and perceptual quality metrics. Fig.~\ref{fig:quali_flow} corroborates these results, with structural artifacts of baseline reconstructions visibly suppressed. Extended evaluations on diffusion model backbones (Stable Diffusion 1.5) are provided in Appendix~\ref{app:flint_diffusion}.

\paragraph{Computational cost and performance.} \label{par:exp_results_3}
Fig.~\ref{fig:robustness_nfe} evaluates SR~$\times$8 and Gaussian deblurring on 1000 FFHQ samples, varying Stage~2 over $\{4, 9, 19, 29\}$ steps (total budgets $\{25, 30, 40, 50\}$). With as few as 30 NFEs, FLINT surpasses nearly all baselines at the full 50 NFE budget on both PSNR and LPIPS. This advantage stems from the proximal initialization. Stage~1 requires only a small number of optimization iterations at low cost enabled by Sec.~\ref{subsec:preliminary_3}, which reduces the computational burden of Stage~2 by positioning the initial noise in a favorable region of the source space, requiring only a short guided trajectory to reach the subspace consistent with the measurement. Extended runtime comparisons are provided in Appendix~\ref{app:add_exp_5}.

\paragraph{Diffusion model.} \label{par:exp_results_4}
To assess applicability beyond flow matching, we further evaluate FLINT on Stable Diffusion 1.5 ($512^2$, 200 NFEs) against PSLD~\cite{psld}, ReSample~\cite{resample}, DDPG~\cite{ddpg}, and TReg~\cite{treg}. As shown in Table~\ref{tab:diffusion_tab}, FLINT achieves the best PSNR, FID, and LPIPS on SR$\times$4 and motion deblurring, as well as the best SSIM on the latter task. Detailed settings and qualitative comparisons are provided in Appendix~\ref{app:flint_diffusion}, with baseline configurations in Appendix~\ref{app:implementation_2}.
\subsection{Ablation Studies} \label{subsec:ablation_studies}
\begin{figure*}[t]
  \centering
  \includegraphics[width=0.9\textwidth]{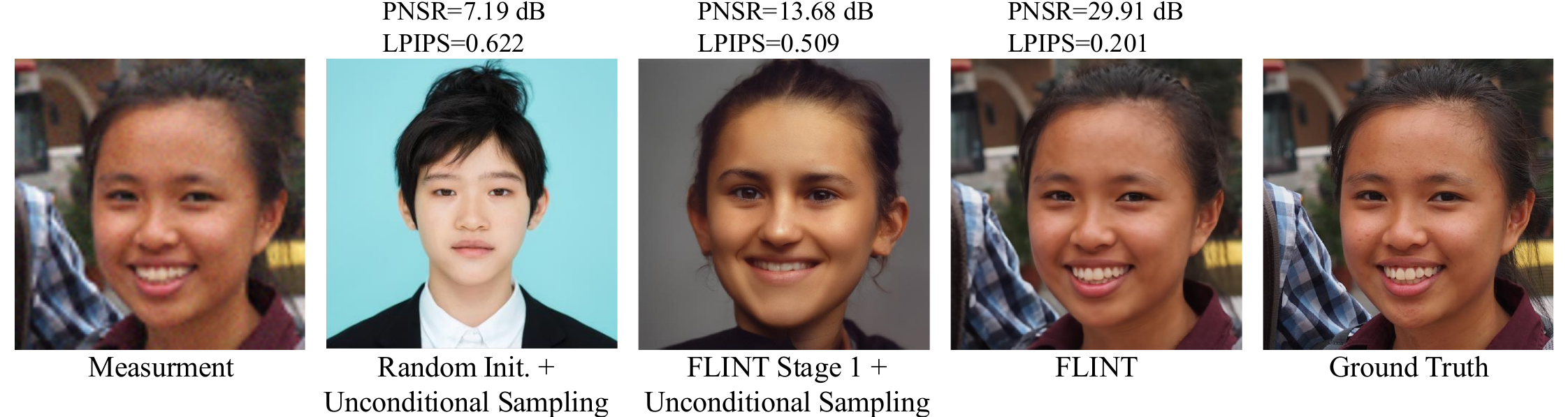}
  \caption{
    \textbf{Initialization quality} on SR$\times$8. Unconditional sampling from the Stage~1 output (third column) demonstrates that FLINT's optimized noise recovers faithful coarse structures, contrasting with the unrelated results from random initialization (second column). Starting from this informed point, Stage~2 achieves high-fidelity reconstruction. Metrics are averaged over 100 FFHQ samples.
  }
  \label{fig:initialization_quality}
\end{figure*}

\paragraph{Effectiveness of stage 1: MCPO and SWDC.}
Table~\ref{tab:ablation_stage1} isolates the contribution of each component within Stage~1, the Manifold-Constrained Proximal Objective (MCPO) and Spectrally Weighted Data Consistency (SWDC). Comparing Conf.~A (random) against Conf.~C (full proximal) confirms that the posterior sampling started from proximal-initialized noise yields substantial gains across all metrics. Figure~\ref{fig:initialization_quality} visually corroborates this: unconditional sampling from the proximal noise recovers a coarse structure consistent with the measurement, whereas the same procedure from a random Gaussian draw produces an entirely unrelated image. Incorporating the SWDC (B to C) improves reconstruction quality by suppressing gradients from unreliable high-frequency components during the optimization, ensuring stable convergence.

\paragraph{Sensitivity to the bounded stochasticity schedule (BSS).}
Table~\ref{tab:ablation_schedule} ablates the stochasticity schedule along two axes: the magnitude $\lambda_s$ and the temporal shape of $\eta(t)$, comparing our bell-shaped schedule (Eq.~\ref{eq:eta_schedule}) against a front-loaded alternative $\eta_{\text{mono}}(t) = \lambda_s \sqrt{1 - (1-\sigma_t)^2}\, \sigma_t$ that decreases monotonically over $t : 1 \to 0$ and peaks at $t=1$. Two key trends emerge. (i) Magnitude: While Sec.~\ref{sec:analysis} warns against excessive stochasticity (large $\lambda_s \ge 1.3$) due to the amplified stochastic departure, suppressing it ($\lambda_s \to 0.3$) limits the FLINT's ability to perform stochastic exploration concentrated at middle timesteps, resulting in suboptimal performance. The moderate setting ($\lambda_s = 0.8$) achieves the optimal balance, bounded to preserve the low-energy proximal state yet active enough to achieve a high-fidelity reconstruction. (ii) Shape: at matched $\lambda_s$, the bell-shaped schedule generally yields superior performance, particularly as $\lambda_s$ increases, as front loading disrupts the proximal noise precisely when coarse structure is being established.

\begin{table*}[t!]
    \centering
    \caption{
    \textbf{Ablation studies} on Gaussian deblurring (FFHQ, 100 samples). (a) Stage 1: Components include MCPO and SWDC. (b) Stage 2: Performance of the BSS profile across $\lambda_s$.
    }
    \resizebox{0.85\textwidth}{!}{
    \renewcommand{\arraystretch}{1.3}
    \setlength{\tabcolsep}{0.1em}
    \begin{subtable}[c]{0.48\textwidth}
        \centering
        \phantomcaption 
        \label{tab:ablation_stage1}
        \vspace{-0.3em}
        \resizebox{\textwidth}{!}{
            \begin{tabular}{c|cc|cccc}
            \toprule
            Conf. & \makecell{MCPO} & \makecell{SWDC} & PSNR$\uparrow$ & SSIM$\uparrow$ & KID$\downarrow$ & LPIPS$\downarrow$ \\ 
            \midrule
            A & \xmarkg & -- & 28.56 & 0.641 & 0.015 & 0.327 \\
            B & \cmarkg & \xmarkg & 29.60 & 0.744 & 0.008 & 0.218 \\
            C & \cmarkg & \cmarkg & 30.11 & 0.788 & 0.005 & 0.192 \\
            \bottomrule
            \end{tabular}
        }
    \end{subtable}
    \hfill
    \renewcommand{\arraystretch}{0.6}
    \setlength{\tabcolsep}{0.4em}
    \begin{subtable}[c]{0.5\textwidth}
        \centering
        \phantomcaption 
        \label{tab:ablation_schedule}
        \vspace{-0.3em}
        \resizebox{\textwidth}{!}{
            \begin{tabular}{cc|cccc}
            \toprule
            Schedule & $\lambda_s$ & PSNR$\uparrow$ & SSIM$\uparrow$ & KID$\downarrow$ & LPIPS$\downarrow$ \\
            \midrule
            \multirow{4}{*}{front-loaded} 
                & 0.3 & 29.60 & 0.769 & 0.022 & 0.193 \\
                & 0.8 & 29.19 & 0.727 & 0.008 & 0.230 \\
                & 1.3 & 28.44 & 0.639 & 0.018 & 0.438 \\
                & 1.8 & 28.32 & 0.629 & 0.027 & 0.499 \\
            \midrule
            \multirow{4}{*}{\makecell{Bell-shaped}}
                & 0.3 & 29.59 & 0.771 & 0.031 & 0.208 \\
                & 0.8 & 30.11 & 0.788 & 0.005 & 0.192 \\
                & 1.3 & 29.88 & 0.784 & 0.024 & 0.305 \\
                & 1.8 & 29.62 & 0.768 & 0.026 & 0.328 \\
            \bottomrule
            \end{tabular}
        }
    \end{subtable}
    }
\end{table*}

\subsection{Discussion} \label{subsec:discussion}
\begin{wrapfigure}{r}{0.5\linewidth}
    \vspace{-1.5em}
    \centering
    \includegraphics[width=\linewidth]{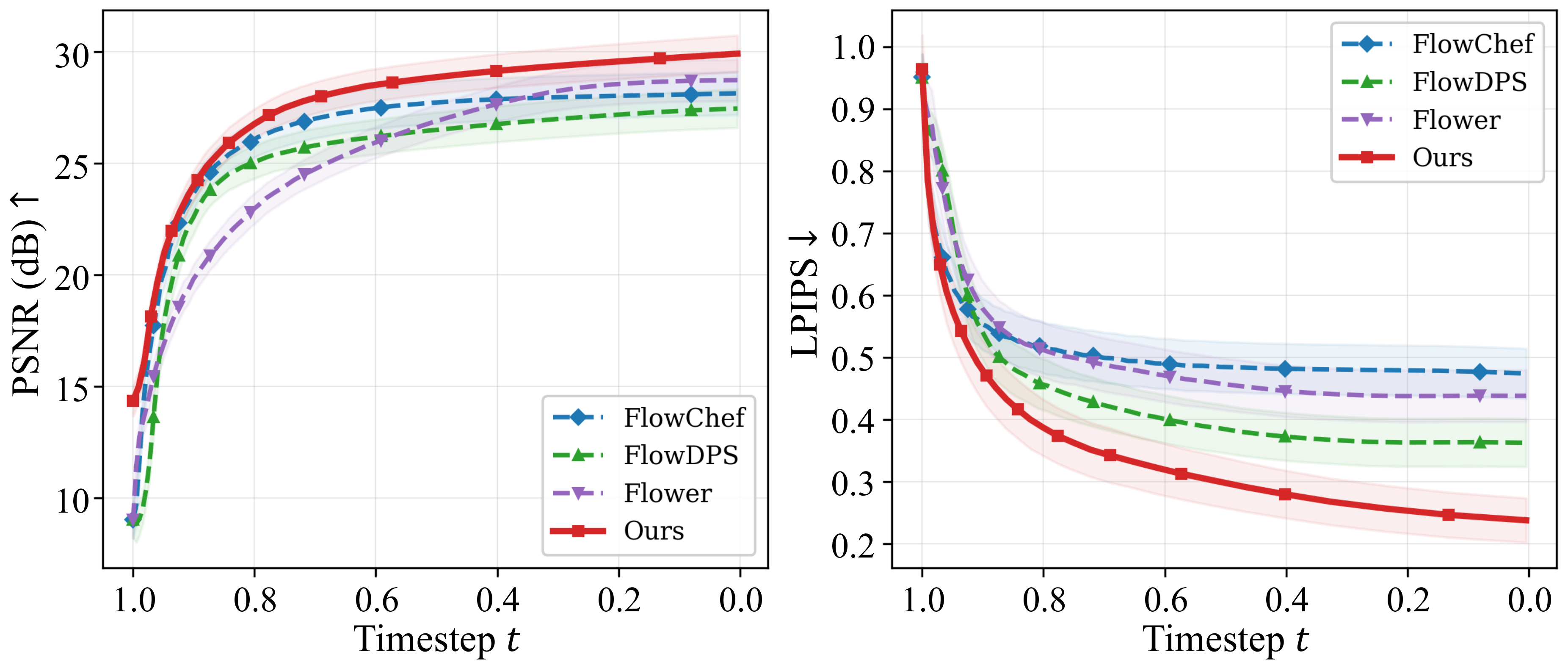}
    \vspace{-1.5em}
    \caption{
    \textbf{Quantitative comparison of the sampling process} via Tweedie's estimates.
    }
    \label{fig:convergence}
\end{wrapfigure}
The theoretical and empirical findings in Sec.~\ref{sec:analysis} predict that proximal initialization reduces the initialization gap, while bounded stochasticity preserves this advantage throughout guided sampling. Fig.~\ref{fig:convergence} corroborates this prediction by evaluating the reconstruction quality of Tweedie's estimate along the sampling trajectory (visualizations in Appendix~\ref{app:add_exp_2}). FLINT enters the measurement-consistent subspace earlier than baselines and sustains higher reconstruction quality through convergence. The Appendix further provides extensive analyses on the effect of the cumulant regularizer (Appendix~\ref{app:add_exp_7}), comparisons with alternative initialization strategies including forward noising initialization~\cite{meng2022sdedit} and flow inversion in both deterministic~\cite{deng2025fireflow} and stochastic variants~\cite{rout2025semantic} (Appendix~\ref{app:add_exp_3}), the effect of prolonged source space optimization (Appendix~\ref{app:add_exp_4}), and a posterior variance analysis confirming the absence of mode collapse (Appendix~\ref{app:add_disc_5}).

\section{Conclusion}
\label{sec:conclusion}
We introduce FLINT, a two-stage framework that bridges source-space and trajectory-space methods by exploiting their coupling, offering a new design principle for flow-based inverse solvers. 
Through a unified stability analysis, we motivate coupling proximal source initialization with bounded stochasticity to preserve the initialization advantage during posterior sampling.
FLINT employs a manifold-constrained proximal objective followed by a strategic stochasticity schedule, consistently improving reconstruction quality across diverse inverse problems at a competitive sampling cost. 

{\bf Limitation.} FLINT inherits the limitations of the underlying pretrained generative models, including training data biases and reduced capacity for out-of-distribution images. Reconstruction quality may degrade on scenes with high structural complexity ($e.g.$, multiple objects or instances), where the increased ill-posedness makes it difficult to simultaneously satisfy data fidelity and prior consistency.

\section*{Acknowledgement}
This work was supported by Institute of Information \& communications Technology Planning \& Evaluation (IITP) grant funded by the Korea government(MSIT) [NO.RS-2021-II211343, Artificial Intelligence Graduate School Program (Seoul National University)], the National Research Foundation of Korea(NRF) grants funded by the Korea government(MSIT) (No. RS-2026-25585456, No. RS-2025-02263628), and Samsung Electronics Co., Ltd. (IO251217-14748-01).

{\small
 \bibliographystyle{plain}
 \bibliography{neurips_2026}
}

\appendix

\renewcommand{\thefigure}{S\arabic{figure}}
\renewcommand{\thetable}{S\arabic{table}}
\renewcommand{\thesection}{S\arabic{section}}
\renewcommand{\theequation}{S\arabic{equation}}

\clearpage
\setcounter{section}{0}
\setcounter{figure}{0}
\setcounter{table}{0}
\setcounter{theorem}{0}
\setcounter{remark}{0}
\renewcommand{\theHremark}{S\arabic{remark}} 
\renewcommand{\theHequation}{S\arabic{equation}} 
\setcounter{equation}{0}
\setcounter{proposition}{0}

\section*{\LARGE{Appendix}}

\appendix

\section{Theoretical Analysis: Assumptions, Propositions, and Proofs} \label{app:theory}

\subsection{Posterior Energy Bound of Principled Initialization} \label{app:subsec_energy_advantage}
This appendix formalizes the interpretation introduced in Remark~\ref{remark:principled_init} (Section~\ref{sec:analysis}): when the reference point $\bm{x}_1^*$ in Proposition~\ref{prop:error_decomposition} is taken to be the posterior energy minimizer, source-space proximity controls a likelihood energy upper bound, while a random Gaussian draw typically lies farther away in source space. 
A principled initialization refers to an initial noise placed deliberately close to the source-space solution, as opposed to a random draw from the prior. We define the posterior energy as the negative log-posterior up to a constant:
\begin{equation}
    \label{eq:posterior_energy}
    E(\bm{x}_1) \;:=\; -\log p(\bm{x}_1 \mid \bm{y}) + C \;=\;
    \underbrace{\frac{1}{2\sigma_y^2}\|\bm{y} - \mathcal{A}(\Phi(\bm{x}_1))\|_2^2}_{E_{\mathrm{like}}(\bm{x}_1)}
    \;+\;
    \underbrace{\frac{1}{2}\|\bm{x}_1\|_2^2}_{E_{\mathrm{prior}}(\bm{x}_1)},
\end{equation}
and choose a source-space solution $\bm{x}_1^* \in \arg\min_{\bm{x}_1}\, E(\bm{x}_1)$.
Throughout the Section~\ref{app:subsec_energy_advantage}, $\bm{n}$ denotes the reference residual, defined by $\bm{n} \;\coloneqq\; \bm{y}-\mathcal{A}(\Phi(\bm{x}_1^*))$.

We state the assumptions underlying Proposition~\ref{prop:energy_advantage}.
\begin{assumption}[Lipschitz generative map]
\label{assump:lipschitz}
    The generative map $\Phi \colon \mathbb{R}^d \to \mathbb{R}^d$ is
    $L$-Lipschitz continuous: for all $\bm{x}_1, \bm{x}_1' \in \mathbb{R}^d$,
    \begin{equation}
        \|\Phi(\bm{x}_1) - \Phi(\bm{x}_1')\|_2
        \leq L\,\|\bm{x}_1 - \bm{x}_1'\|_2.
    \end{equation}
\end{assumption}

\begin{assumption}[Lipschitz forward operator]
\label{assump:lipschitz_A}
    The forward operator $\mathcal{A}$ is $L_A$-Lipschitz continuous in a neighbourhood
    $\mathcal{U}$ of $\Phi(\bm{x}_1^*)$: for all $\bm{u}, \bm{v} \in \mathcal{U}$,
    \begin{equation}
        \|\mathcal{A}(\bm{u}) - \mathcal{A}(\bm{v})\|_2 \leq L_A\,\|\bm{u} - \bm{v}\|_2.
    \end{equation}
    When $\mathcal{A}$ is linear, $L_A = \|\mathcal{A}\|_{\mathrm{op}}$ (the largest singular value of $\mathcal{A}$).
    For nonlinear operators, a finite local Lipschitz constant can be chosen on a sufficiently small neighbourhood whenever $\mathcal{A}$ is continuously differentiable.
\end{assumption}

\begin{assumption}[Typical-set norm of a source-space solution]
\label{assump:x1_norm}
    The source-space solution $\bm{x}_1^*$ satisfies
    \begin{equation}
        \|\bm{x}_1^*\|_2 = \Theta(\sqrt{d}).
    \end{equation}
    This holds when $\bm{x}_1^* \approx \Phi^{-1}(\bm{x}_0^*)$ and the pretrained flow model transports $\mathcal{N}(\bm{0}, \bm{I}_d)$ to the data distribution, since Gaussian concentration implies $\|\bm{z}\|_2 \approx \sqrt{d}$ with high probability for $\bm{z} \sim \mathcal{N}(\bm{0}, \bm{I}_d)$.
\end{assumption}

\begin{proposition}[Posterior energy bound]
\label{prop:energy_advantage}
Let Assumptions~\ref{assump:lipschitz}--\ref{assump:x1_norm} hold. 
Let $\tilde{\bm{x}}_1 \in \mathbb{R}^d$ satisfy 
$\|\tilde{\bm{x}}_1 - \bm{x}_1^*\|_2 \leq \delta$. Then:
\begin{enumerate}[label=(\roman*)]
    \item \textbf{Likelihood energy upper bound:}
    \begin{equation}
        E_{\mathrm{like}}(\tilde{\bm{x}}_1)
        \;\leq\;
        \frac{\bigl(\|\bm{n}\|_2 + L_A \cdot L \cdot \delta\bigr)^2}{2\sigma_y^2}.
    \end{equation}
    \item \textbf{Prior energy near-equality:}
    Both $\tilde{\bm{x}}_1$ (constructed via SLERP with $\|\tilde{\bm{x}}_1\|_2 \approx \sqrt{d}$) and $\bm{x}_1^{\mathrm{rand}} \sim \mathcal{N}(\bm{0}, \bm{I}_d)$ satisfy
    \begin{equation}
        E_{\mathrm{prior}}(\tilde{\bm{x}}_1) 
        \;\approx\; 
        E_{\mathrm{prior}}(\bm{x}_1^{\mathrm{rand}}) 
        \;\approx\; \frac{d}{2}.
    \end{equation}
    \item \textbf{Concentration of random initialization:}
    For $\bm{x}_1^{\mathrm{rand}} \sim \mathcal{N}(\bm{0}, \bm{I}_d)$,
    \begin{equation}
        \|\bm{x}_1^{\mathrm{rand}} - \bm{x}_1^*\|_2 = \Theta(\sqrt{d})
        \quad \text{with high probability.}
    \end{equation}
\end{enumerate}
Since the posterior energy is the sum of the likelihood and prior terms, Claims~(i) and~(ii) together give a proximity-dependent upper bound on the posteirior energy.
\end{proposition}

\begin{proof}
We now prove the three claims of Proposition~\ref{prop:energy_advantage}.

\textbf{Part (i): Upper bound on $E_{\mathrm{like}}(\tilde{\bm{x}}_1)$.}
Since $\bm{y} = A(\bm{x}_0^*) + \bm{n}$ with $\bm{x}_0^* = \Phi(\bm{x}_1^*)$, we have
\begin{align}
\bigl\|\bm{y} - A(\Phi(\tilde{\bm{x}}_1))\bigr\|_2
&= \bigl\|A(\Phi(\bm{x}_1^*)) + \bm{n} - A(\Phi(\tilde{\bm{x}}_1))\bigr\|_2 \notag \\
&\leq \|\bm{n}\|_2 + \bigl\|A(\Phi(\bm{x}_1^*)) - A(\Phi(\tilde{\bm{x}}_1))\bigr\|_2 \tag{triangle inequality} \\
&\leq \|\bm{n}\|_2 + L_A \bigl\|\Phi(\bm{x}_1^*) - \Phi(\tilde{\bm{x}}_1)\bigr\|_2 \tag{Assumption~\ref{assump:lipschitz_A}} \\
&\leq \|\bm{n}\|_2 + L_A \cdot L\, \|\bm{x}_1^* - \tilde{\bm{x}}_1\|_2 \tag{Assumption~\ref{assump:lipschitz}} \\
&\leq \|\bm{n}\|_2 + L_A \cdot L \cdot \delta. \tag{$\|\tilde{\bm{x}}_1 - \bm{x}_1^*\|_2 \leq \delta$}
\end{align}
Squaring and dividing by $2\sigma_y^2$ gives the stated bound.

\medskip
\noindent\textbf{Part (ii): Prior energy near-equality.}
For $\bm{x}_1^{\mathrm{rand}} \sim \mathcal{N}(\bm{0}, \bm{I}_d)$, the strong law of large numbers gives
$\|\bm{x}_1^{\mathrm{rand}}\|_2^2 = d(1+o(1))$ with high probability, so
$E_{\mathrm{prior}}(\bm{x}_1^{\mathrm{rand}}) \approx d/2$.
For $\tilde{\bm{x}}_1$ constructed via SLERP (Eq.~\ref{eq:slerp}), the interpolation preserves
$\|\tilde{\bm{x}}_1\|_2 \approx \sqrt{d}$ by construction, giving
$E_{\mathrm{prior}}(\tilde{\bm{x}}_1) \approx d/2$.

\medskip
\noindent\textbf{Part (iii): Concentration of $\bm{x}_1^{\mathrm{rand}}$.}
Expand the squared distance:
\begin{equation}
\|\bm{x}_1^{\mathrm{rand}} - \bm{x}_1^*\|_2^2
=
\|\bm{x}_1^{\mathrm{rand}}\|_2^2
- 2\langle \bm{x}_1^{\mathrm{rand}},\, \bm{x}_1^* \rangle
+ \|\bm{x}_1^*\|_2^2.
\end{equation}

\emph{Term 1.} By the strong law of large numbers,
$\|\bm{x}_1^{\mathrm{rand}}\|_2^2 = d(1+o(1))$ with high probability.

\emph{Term 2.} Since $\bm{x}_1^{\mathrm{rand}}$ is independent of $\bm{x}_1^*$,
$\mathbb{E}[\langle \bm{x}_1^{\mathrm{rand}}, \bm{x}_1^* \rangle^2]
= \|\bm{x}_1^*\|_2^2 = \Theta(d)$,
so $|\langle \bm{x}_1^{\mathrm{rand}}, \bm{x}_1^* \rangle|/d \to 0$ in $L^2$, and the cross term is $o(d)$.

\emph{Term 3.} By Assumption~\ref{assump:x1_norm},
$\|\bm{x}_1^*\|_2^2 = \Theta(d)$.

Combining all three terms,
$\|\bm{x}_1^{\mathrm{rand}} - \bm{x}_1^*\|_2^2 = \Theta(d)$, and therefore
$\|\bm{x}_1^{\mathrm{rand}} - \bm{x}_1^*\|_2 = \Theta(\sqrt{d})$ with high probability.
\end{proof}

\begin{remark}[Total energy gap]
\label{remark:energy_gap}
Since Claim~(ii) establishes that $E_{\mathrm{prior}}$ is approximately equal for both initializations,
the total energy gap is driven entirely by the likelihood term:
\begin{equation}
E(\tilde{\bm{x}}_1) - E(\bm{x}_1^{\mathrm{rand}})
\;\approx\;
E_{\mathrm{like}}(\tilde{\bm{x}}_1) - E_{\mathrm{like}}(\bm{x}_1^{\mathrm{rand}}).
\end{equation}
Claim~(i) bounds $E_{\mathrm{like}}(\tilde{\bm{x}}_1)$ by a quantity controlled by $\delta$,
while Claim~(iii) implies that
$\|\bm{x}_1^{\mathrm{rand}} - \bm{x}_1^*\|_2 = \Theta(\sqrt{d}) \gg \delta$,
incurring a substantially larger likelihood energy upper bound.
Consequently, when $\delta \ll \sqrt{d}$ and the same Lipschitz bound applies to both initializations, principled initialization admits a tighter distance-based posterior energy upper bound than a random draw.
\end{remark}

\subsection{Proof of Proposition~\ref{prop:error_decomposition}}
\label{app:subsec_proof_error_decomposition}
In the reverse-time convention~\cite{anderson1982reverse,song2020score},
$t$ decreases from $1$ (noise) to $0$ (data), so $dt<0$.
Set $\tau=1-t$ and $\mathbf{X}_\tau:=\mathbf{x}_{1-\tau}$.
Define the sampling velocity $\bar v_t(\cdot|\mathbf{y}):=-v_t(\cdot|\mathbf{y})$,
retaining the original noise-level index $t$; its value at sampling time
$\tau$ is $\bar v_{1-\tau}$. Then~\eqref{eq:governing_sde} becomes
\begin{equation}
  d\mathbf{X}_\tau
  =
    \Bigl[\bar v_{1-\tau}(\mathbf{X}_\tau|\mathbf{y})
          +\tfrac{g_{1-\tau}^2}{2}\nabla_{\mathbf{X}_\tau}\!\log p_{1-\tau}(\mathbf{X}_\tau|\mathbf{y})
    \Bigr]
  \,d\tau
  + g_{1-\tau}\,d\mathbf{W}_\tau,
  \label{eq:posterior_sde_compact}
\end{equation}
where $\mathbf{W}_\tau$ is a standard Wiener process in increasing sampling time.

\begin{assumption}[One-sided Lipschitz posterior velocity]
\label{assum:onesided_vel}
The posterior velocity $\bar v_t(\cdot|\mathbf{y})$ satisfies a one-sided Lipschitz
condition with a measurable, integrable parameter $m_v:[0,1]\to\mathbb{R}$\textup{:}
for all $\mathbf{x},\mathbf{x}'\in\mathbb{R}^d$ and $t\in[0,1]$,
\[
  \bigl\langle \bar v_t(\mathbf{x}|\mathbf{y})-\bar v_t(\mathbf{x}'|\mathbf{y}),\;
               \mathbf{x}-\mathbf{x}'\bigr\rangle
  \;\le\;
  m_v(t)\,\|\mathbf{x}-\mathbf{x}'\|_2^2 .
\]
The parameter $m_v(t)$ may take any sign; negative values indicate
contraction in the sampling direction.
\end{assumption}

\begin{assumption}[One-sided Lipschitz posterior score]
\label{assum:onesided_score}
The posterior score $\nabla_\mathbf{x}\!\log p_t(\mathbf{x}|\mathbf{y})$ satisfies
a one-sided Lipschitz condition with a measurable, integrable parameter
$m_s:[0,1]\to\mathbb{R}$\textup{:} for all $\mathbf{x},\mathbf{x}'\in\mathbb{R}^d$
and $t\in[0,1]$,
\[
  \bigl\langle \nabla\!\log p_t(\mathbf{x}|\mathbf{y})
              -\nabla\!\log p_t(\mathbf{x}'|\mathbf{y}),\;
              \mathbf{x}-\mathbf{x}'\bigr\rangle
  \;\le\;
  m_s(t)\,\|\mathbf{x}-\mathbf{x}'\|_2^2 .
\]
\end{assumption}

\begin{remark}[Comparison with two-sided Lipschitz]
\label{rem:onesided_vs_lipschitz}
A standard two-sided Lipschitz constant bounds the operator norm of the full Jacobian, which is always nonnegative. The one-sided constant above bounds to the maximum eigenvalue of the symmetric part of the Jacobian, which can be \emph{negative}. In particular, log-concavity of $p_t(\cdot|\mathbf{y})$ allows a nonpositive choice of $m_s(t)$, in which case the score term contributes a nonpositive term to the effective contraction rate.
This can reduce the prefactor in our bound by offsetting expansion from the posterior velocity.
\end{remark}

\paragraph{Setup and notation.}
We work with~\eqref{eq:posterior_sde_compact} for $\tau\in[0,1]$, with $\mathbf{y}$ and the deterministic schedule $g$ fixed.
The total drift is
\[
  b(\mathbf{x},\tau)
  :=
  \bar v_{1-\tau}(\mathbf{x}|\mathbf{y})
  +\frac{g_{1-\tau}^2}{2}\nabla_\mathbf{x}\!\log p_{1-\tau}(\mathbf{x}|\mathbf{y}).
\]
Under Assumptions~\ref{assum:onesided_vel}--\ref{assum:onesided_score},
$b$ inherits a one-sided Lipschitz property with a \emph{time-varying}
parameter:
\begin{equation}
  \bigl\langle b(\mathbf{x},\tau)-b(\mathbf{x}',\tau),\;
               \mathbf{x}-\mathbf{x}'\bigr\rangle
  \;\le\;
  m(1-\tau)\,\|\mathbf{x}-\mathbf{x}'\|_2^2,
  \qquad
  m(t):=m_v(t)+\tfrac{g_t^2}{2}\,m_s(t).
  \label{eq:mt_onesided}
\end{equation}
Its time integral defines the contraction parameter:
\[
  M
  \;:=\;
  \int_0^1 m(t)\,dt
  \;=\;
  \int_0^1 m_v(t)\,dt
  \;+\;
  \tfrac{1}{2}\int_0^1 g_t^2\,m_s(t)\,dt.
\]
Note that $M$ may be negative, zero, or positive.

\paragraph{Triangle inequality decomposition.}
We decompose the error via two intermediate reference trajectories.
Couple all processes with the same Wiener process $\bar{\mathbf{W}}_t$:
\[
  \|\Phi_g(\tilde{\mathbf{x}}_1)-\mathbf{x}_0^\star\|_2
  \;\le\;
  \underbrace{%
    \|\Phi_g(\tilde{\mathbf{x}}_1)-\Phi_g(\mathbf{x}_1^\star)\|_2
  }_{\text{Term A}}
  +
  \underbrace{%
    \|\Phi_g(\mathbf{x}_1^\star)-\mathbf{x}_0^\star\|_2
  }_{\text{Term B}} .
\]
Squaring and applying $(a+b)^2\le 2a^2+2b^2$ gives
\begin{equation}
  \mathbb{E}\!\left[
    \|\Phi_g(\tilde{\mathbf{x}}_1)-\mathbf{x}_0^\star\|_2^2
  \right]
  \;\le\;
  2\,\mathbb{E}\bigl[\|\Phi_g(\tilde{\mathbf{x}}_1)-\Phi_g(\mathbf{x}_1^\star)\|_2^2\bigr]
  + 2\,\mathbb{E}\bigl[\|\Phi_g(\mathbf{x}_1^\star)-\mathbf{x}_0^\star\|_2^2\bigr].
  \label{eq:triangle_squared}
\end{equation}

\paragraph{Bounding Term A (initialization sensitivity).}
Drive both trajectories $\Phi_g(\tilde{\mathbf{x}}_1)$ and $\Phi_g(\mathbf{x}_1^\star)$
with the same Wiener process. Define the pathwise error
$\mathbf{e}_\tau:=\mathbf{X}_\tau^{(\tilde{\mathbf{x}}_1)}-\mathbf{X}_\tau^{(\mathbf{x}_1^\star)}$.
Subtracting the two SDEs, the diffusion terms cancel exactly:
\[
  d\mathbf{e}_\tau
  =
  \bigl[b(\mathbf{X}_\tau^{(\tilde{\mathbf{x}}_1)},\tau)
        -b(\mathbf{X}_\tau^{(\mathbf{x}_1^\star)},\tau)\bigr]d\tau,
\]
so $\mathbf{e}_\tau$ satisfies a deterministic ODE. 
Differentiating the squared norm and applying~\eqref{eq:mt_onesided} gives,
for almost every $\tau$, almost surely,
\[
  \frac{d}{d\tau}\|\mathbf{e}_\tau\|_2^2
  \;=\;
  2\bigl\langle\mathbf{e}_\tau,\;
    b(\mathbf{X}_\tau^{(\tilde{\mathbf{x}}_1)},\tau)
    -b(\mathbf{X}_\tau^{(\mathbf{x}_1^\star)},\tau)\bigr\rangle
  \;\le\;
  2m(1-\tau)\,\|\mathbf{e}_\tau\|_2^2 .
\]
The initial condition is $\|\mathbf{e}_0\|_2^2 = \|\tilde{\mathbf{x}}_1
-\mathbf{x}_1^\star\|_2^2 \le \delta^2$. Gronwall's inequality~\cite{walter1998ordinary}
gives the pathwise bound
\[
  \|\mathbf{e}_1\|_2^2
  \;\le\;
  \exp\!\Bigl(\int_0^1 2m(1-\tau)\,d\tau\Bigr)\delta^2
  \;=\;
  e^{2M}\,\delta^2 .
\]
Taking expectations yields
\begin{equation}
  \mathbb{E}\bigl[\|\Phi_g(\tilde{\mathbf{x}}_1)-\Phi_g(\mathbf{x}_1^\star)\|_2^2\bigr]
  \;\le\;
  e^{2M}\,\delta^2 .
  \label{eq:termA_onesided}
\end{equation}

\paragraph{Bounding Term B (stochastic departure accumulation).}
Define $\boldsymbol{\xi}_\tau:=\mathbf{X}_\tau^{(\eta)}-\mathbf{X}_\tau^{(0)}$, the
difference between
\begin{itemize}
  \item $\mathbf{X}_\tau^{(\eta)}$: trajectory of~\eqref{eq:posterior_sde_compact}
        initialized at $\mathbf{X}_0^{(\eta)}=\mathbf{x}_1^\star$, and
  \item $\mathbf{X}_\tau^{(0)}$: drift ODE trajectory,
        $d\mathbf{X}_\tau^{(0)}=b(\mathbf{X}_\tau^{(0)},\tau)\,d\tau$,
        initialized at $\mathbf{X}_0^{(0)}=\mathbf{x}_1^\star$,
        so that $\mathbf{X}_1^{(0)}=\mathbf{x}_0^\star$.
\end{itemize}
Here $(0)$ denotes removal of only the Brownian term, with the same
schedule $g$ retained in $b$.
Both trajectories share the same total drift $b$, so
\[
  d\boldsymbol{\xi}_\tau
  =
  \bigl[b(\mathbf{X}_\tau^{(\eta)},\tau)-b(\mathbf{X}_\tau^{(0)},\tau)\bigr]d\tau
  + g_{1-\tau}\,d\mathbf{W}_\tau,
  \qquad
  \boldsymbol{\xi}_0=\mathbf{0}.
\]
Applying It\^{o}'s lemma to $\|\boldsymbol{\xi}_\tau\|_2^2$ gives
\[
  d\|\boldsymbol{\xi}_\tau\|_2^2
  =
  2\langle\boldsymbol{\xi}_\tau,d\boldsymbol{\xi}_\tau\rangle
  + d\,g_{1-\tau}^2\,d\tau.
\]
Using the one-sided bound~\eqref{eq:mt_onesided}, a standard localization and
integrating-factor argument, followed by Fatou's lemma, yields Gronwall's
estimate in the original time variable:
\begin{equation}
  \mathbb{E}\|\boldsymbol{\xi}_1\|_2^2
  \;\le\;
  d\int_0^1 g_t^2\,
    \exp\!\Bigl(2\int_0^t m(u)\,du\Bigr)\,dt
  \;\le\; e^{2M_+}\,d\cdot G,
  \label{eq:xi_gronwall_onesided}
\end{equation}
where $M_+ := \max\bigl(0,\;\sup_{s\in[0,1]}\int_0^s m(u)\,du\bigr)$ controls
the worst-case partial accumulation along the trajectory. Therefore,
\begin{equation}
  \mathbb{E}\bigl[\|\Phi_g(\mathbf{x}_1^\star)-\mathbf{x}_0^\star\|_2^2\bigr]
  \;\le\;
  e^{2M_+}\,d\cdot G .
  \label{eq:termB_onesided}
\end{equation}

\paragraph{Combining the bounds.}
Since $M\le M_+$, substituting~\eqref{eq:termA_onesided} and~\eqref{eq:termB_onesided}
into~\eqref{eq:triangle_squared} gives
\[
  \mathbb{E}\!\left[
    \|\Phi_g(\tilde{\mathbf{x}}_1)-\mathbf{x}_0^\star\|_2^2
  \right]
  \;\le\;
  2\,e^{2M}\,\delta^2
  + 2\,e^{2M_+}\,d\cdot G
  \;\le\;
  2\,e^{2M_+}
  \bigl(\delta^2 + d\cdot G\bigr).
\]
This completes the proof.

\section{Additional Empirical Validation: Erosion by Excessive Stochasticity} \label{app:more_analysis}
\begin{figure}[!htbp]
    \centering
    \includegraphics[width=0.65\linewidth]{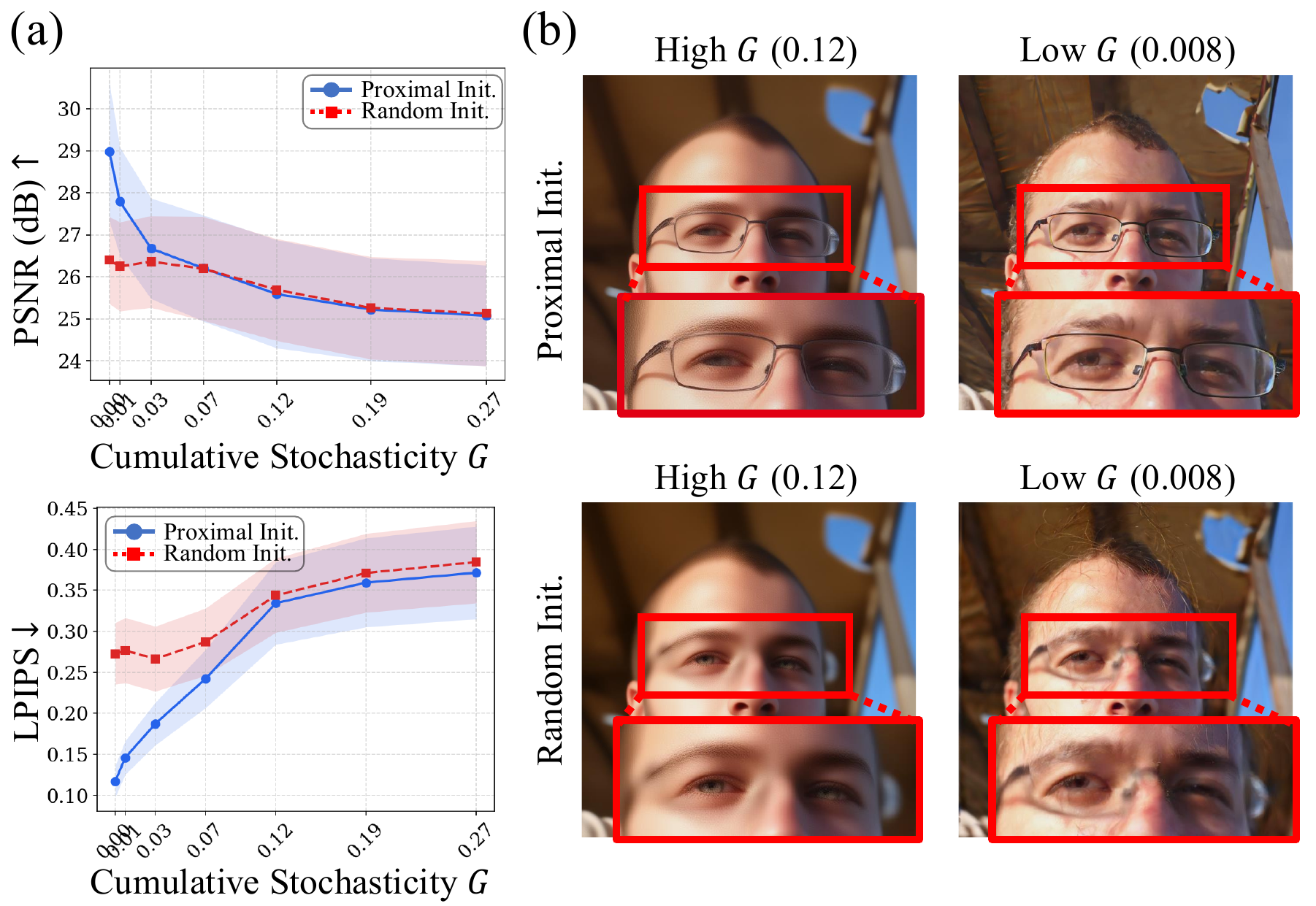}
    \caption{
    Additional empirical validation for erosion by excessive stochasticity. (a) PSNR and LPIPS as a function of the cumulative stochasticity $G$ for proximal initialization ($C_0 = 0.4$) and random initialization ($C_0 = \pi/2$). Proximal initialization outperforms random initialization at low $G$, while the performance gap narrows as $G$ increases. (b) Qualitative comparison across initialization types (proximal vs.\ random, rows) and stochasticity levels (high $G = 0.12$ vs.\ low $G = 0.008$, columns).
    }
    \label{fig:app_analysis_2}
\end{figure}

This section provides additional empirical validation for the erosion phenomenon identified in Section~3, complementing the joint analysis of Figure~\ref{fig:analysis_1}(a) with a controlled experiment that isolates the effect of exessive cumulative stochasticity from the initialization gap. We fix the initialization at either proximal ($C_0 = 0.4$) or random ($C_0 = \pi/2$) and vary the cumulative stochasticity $G$. Under this design, Figure~\ref{fig:app_analysis_2}(a) shows that at low $G$, proximal initialization substantially outperforms random initialization across both PSNR and LPIPS, confirming that a well-chosen starting point confers a meaningful reconstruction advantage when cumulative stochasticity is bounded. 
As $G$ grows, the performance gap narrows and eventually vanishes. The direction of this trend agrees with the monotone dependence of the bound in Proposition 3.1 on $G$, although the bound itself does not predict that the gap closes.
Figure~\ref{fig:app_analysis_2}(b) corroborates this qualitatively. At low stochasticity ($G = 0.008$), the proximal initialization produces reconstructions that are perceptually sharper and more faithful to the ground truth than the random initialization, demonstrating that the initialization advantage translates into visible quality differences. At high stochasticity ($G = 0.12$), both initialization types yield reconstructions of comparable and degraded quality, visually confirming that excessive cumulative noise injection overrides any benefit from a principled starting point. Together, these results suggest that bounding $G$ plays a key role in preserving the initialization advantage throughout guided sampling.

\section{Further Details on FLINT Algorithm} \label{app:details_flint}
\subsection{Stage 1: Proximal Source Initialization} \label{app:details_flint_1}
We cast source-space initialization as variational optimization by introducing a
variational distribution $q_\phi(\bm{x}_1) = \mathcal{N}(\bm{x}_1 \mid \boldsymbol{\mu}_\phi, \rho^2 \mathbf{I}_d)$ over the initial noise and minimizing the objective in Eq.~\eqref{eq:source_proximal} of the main paper.
The expectation in Eq.~\eqref{eq:source_proximal} is approximated by a single reparameterized sample per iteration, i.e., $\bm{x}_1 = \boldsymbol{\mu}_\phi + \rho\,\bm{\varepsilon}$, $\bm{\varepsilon} \sim \mathcal{N}(\bm{0}, \mathbf{I}_d)$. We find this sufficient in practice given the smoothness of $\ell_{\mathrm{IR}}$ over the spectrally filtered subspace~\ref{par:method_1_2}.  Gradients $\nabla_\phi \mathcal{J}(\phi)$ are computed via the SDO shortcut combined with the truncated ODE forward pass.

\paragraph{Closed-form KL term.} \label{app:details_flint_1_2}
Since both $q_\phi$ and the prior $p(\bm{x}_1) = \mathcal{N}(\bm{0},\mathbf{I}_d)$ are Gaussian, the KL divergence admits a closed form. For two $d$-dimensional
Gaussians with diagonal covariance,
\begin{equation}
    \mathrm{KL}\!\bigl(\mathcal{N}(\boldsymbol{\mu},\rho^2\mathbf{I}_d)
    \;\|\;\mathcal{N}(\bm{0},\mathbf{I}_d)\bigr)
    = \frac{1}{2}
      \Bigl[
        \mathrm{tr}(\rho^2 \mathbf{I}_d)
        + \boldsymbol{\mu}^\top \mathbf{I}_d^{-1}\boldsymbol{\mu}
        - d
        + \log\frac{\det(\mathbf{I}_d)}{\det(\rho^2\mathbf{I}_d)}
      \Bigr],
\end{equation}
which, upon substituting $\det(\rho^2 \mathbf{I}_d) = \rho^{2d}$ and
$\log \det(\mathbf{I}_d) = 0$, simplifies to
\begin{equation}
    \mathrm{KL}\!\bigl(q_\phi \,\|\, p\bigr)
    = \frac{1}{2}
      \Bigl(
        \|\boldsymbol{\mu}_\phi\|_2^2
        + d\rho^2
        - d
        - d\log\rho^2
      \Bigr).
\end{equation}
This term simultaneously penalizes the magnitude of $\boldsymbol{\mu}_\phi$ (through $\|\boldsymbol{\mu}_\phi\|_2^2$) and constrains the variance $\rho^2$ toward unity, regularizing $q_\phi$ toward $p(\bm{x}_1)$ in second-order statistics.

\paragraph{Closed-form cumulant regularizer.} \label{app:details_flint_1_3}
The KL term alone cannot preclude higher-order departures from the Gaussian prior, for instance, a heavy-tailed or skewed coordinate distribution that shares the first two moments of $\mathcal{N}(0,1)$. To address this, we include the cumulant regularizer $\mathcal{R}(\boldsymbol{\mu}_\phi)$, which penalizes non-zero standardized skewness and excess kurtosis of the per-channel coordinate distribution of $\boldsymbol{\mu}_\phi$. Treating the $HW$ spatial coordinates of channel $c$ as an empirical sample, define the per-channel empirical mean and variance:
\begin{equation}
    \bar{\mu}_c = \frac{1}{HW}\sum_{i=1}^{HW}(\mu_\phi)_{c,i},
    \qquad
    \hat{s}_c^{\,2} = \frac{1}{HW}\sum_{i=1}^{HW}\bigl[(\mu_\phi)_{c,i} - \bar{\mu}_c\bigr]^2.
\end{equation}
The per-channel standardized skewness $\gamma_{1,c}$ and excess kurtosis $\gamma_{2,c}$ are then
\begin{align}
    \gamma_{1,c}(\boldsymbol{\mu}_\phi)
    &= \frac{1}{HW}\sum_{i=1}^{HW}
       \left(\frac{(\mu_\phi)_{c,i} - \bar{\mu}_c}{\hat{s}_c}\right)^{\!3},
    \label{eq:skewness}\\[4pt]
    \gamma_{2,c}(\boldsymbol{\mu}_\phi)
    &= \frac{1}{HW}\sum_{i=1}^{HW}
       \left(\frac{(\mu_\phi)_{c,i} - \bar{\mu}_c}{\hat{s}_c}\right)^{\!4}
       - 3,
    \label{eq:kurtosis}
\end{align}
where both quantities vanish for a perfectly Gaussian coordinate distribution. The regularizer averages these penalties over the $C$ channels:
\begin{equation}
    \mathcal{R}(\boldsymbol{\mu}_\phi)
    = \frac{1}{C}\sum_{c=1}^{C}
      \left[\frac{1}{12}\,\gamma_{1,c}^2(\boldsymbol{\mu}_\phi)
      + \frac{1}{48}\,\gamma_{2,c}^2(\boldsymbol{\mu}_\phi)\right].
    \label{eq:cumulant_reg}
\end{equation}
The coefficients $\tfrac{1}{12}$ and $\tfrac{1}{48}$ are not chosen ad hoc but follow from the Edgeworth expansion of $\log p$ around $\mathcal{N}(0,1)$: at third and fourth cumulant order, the expansion yields contributions of $\tfrac{\kappa_3^2}{12}$ and $\tfrac{\kappa_4^2}{48}$ to $\mathrm{KL}(p\|\mathcal{N}(0,1))$~\cite{mccullagh2018tensor}. Consequently, $\mathcal{R}$ constitutes a second-order approximation to the KL divergence between the empirical marginal coordinate distribution of each channel and $\mathcal{N}(0,1)$. Together, the KL term controls the norm of $\boldsymbol{\mu}_\phi$, while $\mathcal{R}$ suppresses skewness and excess kurtosis, jointly reducing the risk of the optimized noise drifting outside the training support of the pretrained flow model during optimization.

\paragraph{Noise annealing schedule.} \label{app:details_flint_1_1}
The variance $\rho$ is linearly annealed over $K$ optimization iterations:
\begin{equation}
    \rho^{(s)} = \rho_{\max} - \frac{s}{S}\!\left(\rho_{\max} - \rho_{\min}\right),
    \quad s = 0, 1, \ldots, S,
\end{equation}
with $\rho_{\min} = 0.02$.  Large $\rho$ in early iterations smooths the optimization landscape and prevents premature convergence to suboptimal $\boldsymbol{\mu}_\phi$. As $\rho$ decays toward $\rho_{\min}$, gradient estimates concentrate around $\boldsymbol{\mu}_\phi$, enabling fine-grained refinement within the identified low-energy region.

\subsection{Stage 2: Proximity-Preserving Posterior Sampling with Bounded Stochasticity} \label{app:details_flint_2}
At the start of Stage 2, the sample $\tilde{\bm{x}}_1=\bm{\mu}_\phi$ produced by Stage~1 replaces the standard random initialization, positioning the sampler within a low posterior energy region from the outset. Each reverse step proceeds through four sequential operations.

\paragraph{Text-conditional velocity field.}
At each timestep $t$, the pretrained flow model is queried under the text condition $c$ with a fixed classifier-free guidance scale $\lambda$:
\begin{equation}
    v_t(\bm{x}_t)
    = v_\theta(\bm{x}_t, c_\emptyset)
    + \lambda\,\bigl(v_\theta(\bm{x}_t, c) - v_\theta(\bm{x}_t, c_\emptyset)\bigr).
\end{equation}

\paragraph{Decomposition via Tweedie's formula.}
The velocity $v_t(\bm{x}_t)$ simultaneously yields a clean image estimate $\hat{\bm{x}}_{0|t}$ and a source noise estimate $\hat{\bm{x}}_{1|t}$ via the flow-based Tweedie's formula:
\begin{equation}
\label{eq:tweedie_update}
    \hat{\bm{x}}_{0|t} = \bm{x}_t - \sigma_t\, v_t(\bm{x}_t),
    \qquad
    \hat{\bm{x}}_{1|t} = \bm{x}_t + (1 - \sigma_t)\, v_t(\bm{x}_t),
\end{equation}
where $\sigma_t$ denotes the noise schedule.

\paragraph{Measurement-consistency correction.}
\begin{equation}
\label{eq:dc_update}
    \tilde{\bm{x}}_{0|t}(\bm{y}) = \underset{\bm{x}}{\arg\min}\;
       \mathcal{L}\!\bigl(\mathcal{A}(\bm{x}),\,\bm{y}\bigr)
    = \hat{\bm{x}}_{0|t}
    - \beta(t)\,\nabla_{\hat{\bm{x}}_{0|t}}
      \mathcal{L}\!\bigl(\mathcal{A}(\hat{\bm{x}}_{0|t}),\,\bm{y}\bigr),
\end{equation}
where $\mathcal{L}$ measures the discrepancy between the forward projection of the current estimate and the observed measurement, and $\beta(t)$ 
is a time-varying data consistency guidance scale. The loss $\mathcal{L}$ in Eq.~\eqref{eq:dc_update} is instantiated as a weighted interpolation between a Back-Projection (BP) objective and a Least-Squares (LS) objective~\citep{ddpg}:
\begin{equation}
\label{eq:hybrid_dc}
    \mathcal{L}\!\bigl(\mathcal{A}(\hat{\bm{x}}_{0|t}),\,\bm{y}\bigr)
    = (1 - \omega_t)\,
      \bigl\|
        \mathcal{A}^\dagger(\bm{y})
        - \mathcal{A}^\dagger\!\bigl(\mathcal{A}(\hat{\bm{x}}_{0|t})\bigr)
      \bigr\|_2
    + \omega_t\,
      \bigl\|\bm{y} - \mathcal{A}(\hat{\bm{x}}_{0|t})\bigr\|_2,
\end{equation}
where $\mathcal{A}^\dagger$ is the pseudo-inverse of $\mathcal{A}$. The interpolation weight $\omega_t = (1 - \sigma_\omega(t))^\phi$ with $\phi = 0.8$ transitions from a BP dominated correction at high noise levels to an LS dominated correction as $t \to 0$.  For flow matching models, $\sigma_\omega(t) = \sigma_t$. For diffusion models, $\sigma_\omega(t) = 1 - \bar{\alpha}_t$.

\paragraph{Euler update with bounded stochasticity.}
To preserve the initialization advantage from Stage~1, the next state is obtained by renoising the corrected clean estimate $\tilde{\bm{x}}_{0|t}(\bm{y})$ to the $(t + \Delta t)$-manifold with stochasticity injected into the source noise component under the bounded schedule $\eta(t)$:
\begin{equation}
    \bm{x}_{t + \Delta t}
    = (1 - \sigma_{t + \Delta t})\,\tilde{\bm{x}}_{0|t}(\bm{y})
      + \sigma_{t + \Delta t}\,\hat{\bm{x}}_{1|t}
      + \eta(t)\,\bm{\epsilon}, \quad \bm{\epsilon} \sim \mathcal{N}(\bm{0}, \mathbf{I}).
\end{equation}
The complete inference procedure is given in Algorithm~\ref{alg:flint}.

\begin{algorithm}[!t]
\caption{Truncated Sampling with SDO Shortcut}\label{alg:sdo_euler}
\begin{algorithmic}[1]
\Require $\bm{x}_1$, $\bm{v}_\theta$, $(c, c_\varnothing)$, $\lambda$, $K$
\State $\bm{x} \gets \bm{x}_1$
\For{$i = 1, \ldots, K-1$}
    \State $\bm{v} \gets \bm{v}_\theta(\bm{x}, c_\varnothing) + \lambda \, (\bm{v}_\theta(\bm{x}, c) - \bm{v}_\theta(\bm{x}, c_\varnothing))$
    \If{$i \ne 1$}
        \State $\bm{v} \gets \mathrm{stop\_grad}(\bm{v})$ \hfill $\triangleright$ SDO shortcut: retain graph only at $i=1$
    \EndIf
    \State $\bm{x} \gets \bm{x} + (\tau_{i+1} - \tau_i)\, \bm{v}$ \hfill $\triangleright$ Euler update; $\tau_i = \tau_i^{(N)}$
\EndFor
\State $t_{\mathrm{stop}} \gets \tau_{K-1}$
\State $\bm{v}_{t_{\mathrm{stop}}} \gets \mathrm{stop\_grad}\!\left(\bm{v}_\theta(\bm{x},\, c_\varnothing) + \lambda\,(\bm{v}_\theta(\bm{x},\, c) - \bm{v}_\theta(\bm{x},\, c_\varnothing))\right)$
\State $\hat{\bm{x}}_0 \gets \bm{x} - t_{\mathrm{stop}} \cdot \bm{v}_{t_{\mathrm{stop}}}$ \hfill $\triangleright$ Tweedie's formula (Eq.~\eqref{eq:truncated_tweedie})
\State \Return $\hat{\bm{x}}_0$
\end{algorithmic}
\end{algorithm}


\section{FLINT in Diffusion Model} \label{app:flint_diffusion}
\subsection{Datasets and Metrics} \label{app:flint_diffusion_1}
Quantitative evaluation is conducted on 1,000 high resolution images from the FFHQ dataset~\cite{ffhq}. Distortion is measured via PSNR and SSIM~\cite{ssim}, and perceptual quality via LPIPS~\cite{lpips} and a patch based FID~\cite{fid} computed on $128\times128$ patches using InceptionV3 features.

\subsection{Baselines} \label{app:flint_diffusion_2}
We evaluate our approach against diffusion based baselines implemented on Stable Diffusion 1.5 ($512^2$, NFE 200), encompassing PSLD~\cite{psld}, ReSample~\cite{resample}, DDPG~\cite{ddpg}, and TReg~\cite{treg}. Hyperparameters for each baseline are provided in ~\ref{app:implementation_2}.

\subsection{Problem Setting}
Diffusion based experiments cover two inverse problems: super-resolution 4$\times$ and motion deblurring with a kernel size of 61 and intensity of 0.5. Across all settings, Gaussian noise with standard deviation of 0.03 (1.5\% of the pixel range) is added to the measurements.

\subsection{Experimental Results for Diffusion Model}
\begin{figure*}[htbp!]
  \centering
  \includegraphics[width=1.0\textwidth]{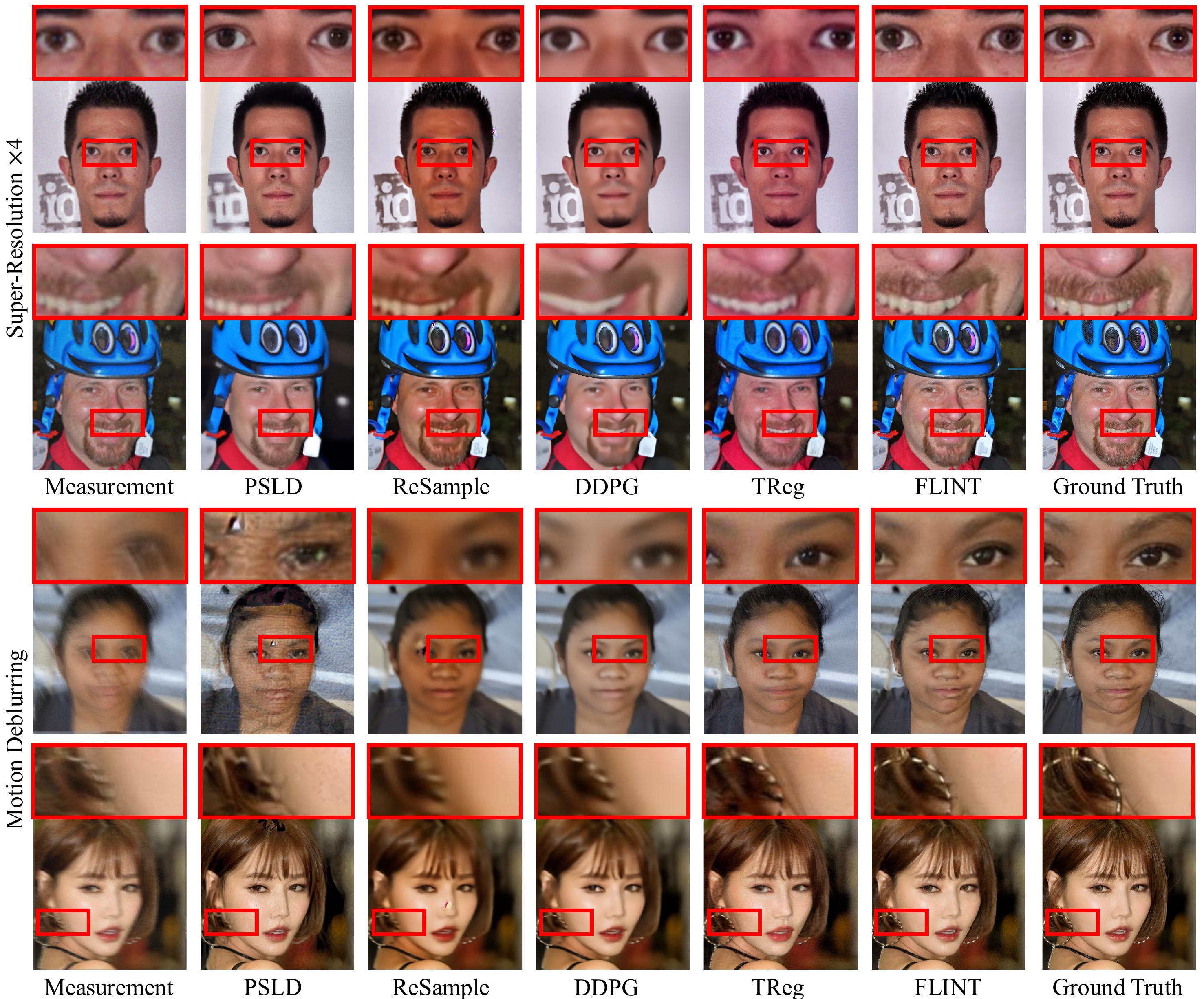}
  \caption{
  Qualitative comparison for FFHQ dataset on super-resolution 4$\times$ and motion deblurring based on diffusion model (SD 1.5). 
  }
  \label{fig:app_quali_diffusion_model}
\end{figure*}

To evaluate the generalizability of FLINT beyond flow matching frameworks, we extend the proposed method to diffusion models and benchmark it against diffusion based baselines on super resolution $\times4$ and motion deblurring. As summarized in Table~\ref{tab:diffusion_tab}, FLINT achieves the best distortion perception trade-off across both tasks, demonstrating that proximal initialization via variational framework and guided posterior sampling with a bounded stochasticity schedule generalize effectively to the diffusion model setting. The qualitative comparisons in Fig.~\ref{fig:app_quali_diffusion_model} corroborate these findings, with FLINT producing reconstructions of higher visual fidelity and structural coherence than competing baselines. Collectively, these results confirm that the core principles of FLINT are not specific to rectified flow priors but transfer robustly across distinct generative mechanisms.

\section{Implementation Details} \label{app:implementation}
\subsection{Implementation Details for FLINT} \label{app:implementation_1}

\subsubsection{Stage 1: Proximal Source Initialization}
\paragraph{Optimization setup.}
The optimization parameters $\boldsymbol{\mu}_\phi$ are optimized using the Adam optimizer. The regularization coefficients are set to $\beta_{\mathrm{KL}} = 0.1$ and $\lambda_{\mathrm{cum}} = 10$ for FFHQ under the flow matching framework, with the learning rate ($\alpha_{\mathrm{lr}}$) set to $0.06$. The noise annealing schedule uses $\sigma_{\max} = 0.5$ and $\sigma_{\min} = 0.02$ throughout all flow matching experiments. Hyperparameters for DIV2K and diffusion model experiments are summarized in Table~\ref{tab:impl_stage1}.

\begin{table*}[!ht]
    \centering
    \renewcommand{\arraystretch}{0.9}
    \setlength{\tabcolsep}{0.6em}
    \caption{Stage 1 hyperparameters across datasets and generative frameworks.}
    \label{tab:impl_stage1}
    \resizebox{0.7\textwidth}{!}{
    \begin{tabular}{l|l|cccc}
    \toprule
    Framework & Dataset & $\beta_{\mathrm{KL}}$ & $\lambda_{\mathrm{cum}}$ & $\sigma_{\max}$ & $\alpha_{\mathrm{lr}}$ \\
    \midrule
    \multirow{2}{*}{Flow Matching (SD3.5-M)}
        & FFHQ  & 0.1  & 10.0 & 0.5 & 0.006 \\
        & DIV2K & 0.01 & 1.0  & 0.5 & 0.01  \\
    \midrule
    Diffusion (SD1.5)
        & FFHQ  & 0.01 & 10.0 & 0.02 & 0.02 \\
    \bottomrule
    \end{tabular}
    }
\end{table*}

\paragraph{Truncated sampling configuration.}
For flow matching experiments, we use $K = 3$ truncated ODE steps per optimization iteration, with $7$ optimization iterations in total, yielding $\mathrm{NFE}_{\mathrm{Stage1}} = K \times 7 = 21$ function evaluations. The remaining $29$ NFEs are allocated to Stage~2 guided posterior sampling, for a total budget of $50$ NFEs. For diffusion model experiments, we use $K = 10$ truncated ODE steps per optimization iteration with $10$ optimization iterations, yielding $\mathrm{NFE}_{\mathrm{Stage1}} = K \times 10 = 100$ function evaluations, with the remaining $100$ NFEs allocated to Stage~2, for a total of $200$ NFEs.

\paragraph{Calibration for spectrally weighted data consistency.}
The Wiener filter weight $W(\omega, t_{\text{stop}})$ in Eq.~\ref{eq:spectral_dc} assumes a pink noise prior $S_{x_0}(\omega) \propto \|\omega\|^{-\alpha}$ for natural images~\cite{simoncelli2001natural}, consistent with the $1/f^2$ power spectrum, with $\alpha = 2.0$ and $t_{\text{stop}} = 0.976$ fixed across all tasks and datasets. While $\alpha$ determines the spectral shape of the prior, it leaves the proportionality constant undetermined, causing the transition frequency between near unity and near zero mask values to shift in an uncontrolled manner. Calibration resolves this ambiguity by rescaling the raw mask to attain a prescribed value $W_{\text{at\_cutoff}}$ at a designated reference frequency $\omega_c$, thereby anchoring the transition between reliable and unreliable frequency components as a half power point. We uniformly adopt $\omega_c = 0.05$, $W_{\text{at\_cutoff}} = 0.5$, and $\alpha = 2.0$ across all tasks and datasets without task specific tuning, indicating that the calibration scheme is robust to variations in degradation type and image domain.

\subsubsection{Stage 2: Proximity-Preserving Posterior Sampling with Bounded Stochasticity}
\paragraph{Stochasticity magnitude.}
The overall magnitude $\lambda_s$ of the bounded stochasticity schedule $\eta(t)$ in Eq.~\ref{eq:eta_schedule} governs the total amount of noise injected throughout the guided sampling trajectory and is likewise task-dependent. A complete specification of the $\lambda_s$ values used across all datasets and inverse problems is provided in Table~\ref{tab:impl_stage2_lambda}.

\paragraph{Data consistency guidance scale.}
The time-varying guidance scale $\beta(t)$ in Eq.~\ref{eq:dc_update} controls the step size of the measurement correction at each reverse timestep and is task-dependent. Specifically, $\beta(t)$ follows a monotonically decreasing schedule interpolating between a minimum value $\beta_{\min}$ and a maximum value $\beta_{\max}$ over the sampling trajectory. The complete set of $(\beta_{\min}, \beta_{\max})$ values used across all datasets and inverse problems is provided in Table~\ref{tab:impl_stage2_beta}.

\begin{table*}[!ht]
    \centering
    \renewcommand{\arraystretch}{0.9}
    \setlength{\tabcolsep}{0.6em}
    \caption{{Stochasticity magnitude $\lambda_s$ across tasks and datasets.}}
    \label{tab:impl_stage2_lambda}
    \resizebox{0.9\textwidth}{!}{
    \begin{tabular}{l|l|cccc}
    \toprule
    Framework & Dataset & SR $\times$8 & SR $\times$12 & Motion Deblurring & Gaussian Deblurring \\
    \midrule
    \multirow{2}{*}{Flow Matching (SD3.5-M)}
        & FFHQ  & 0.9 & 0.9 & 0.6 & 0.8 \\
        & DIV2K & 0.9 & 0.8 & 0.9 & 0.9 \\
    \midrule
    Framework & Dataset & SR $\times$4 & Motion Deblur & & \\
    \midrule
    Diffusion (SD1.5)
        & FFHQ  & 0.3 & 0.3 & & \\
    \bottomrule
    \end{tabular}
    }
\end{table*}

\begin{table*}[!ht]
    \centering
    \renewcommand{\arraystretch}{0.9}
    \setlength{\tabcolsep}{0.6em}
    \caption{{Data consistency guidance scale $(\beta_{\min}, \beta_{\max})$ configurations across tasks and datasets.}}
    \label{tab:impl_stage2_beta}
    \resizebox{0.9\textwidth}{!}{
    \begin{tabular}{l|l|cccc}
    \toprule
    Framework & Dataset & SR $\times$8 & SR $\times$12 & Motion Deblurring & Gaussian Deblurring \\
    \midrule
    \multirow{2}{*}{Flow Matching (SD3.5-M)}
        & FFHQ  & (70, 130) & (30, 130) & (110, 130) & (60, 140) \\
        & DIV2K & (125, 50) & (120, 0) & (450, 250) & (160, 100) \\
    \midrule
    Framework & Dataset & SR $\times$4 & Motion Deblur & & \\
    \midrule
    Diffusion (SD1.5)
        & FFHQ  & (10, 100) & (50, 20) & & \\
    \bottomrule
    \end{tabular}
    }
\end{table*}

\subsection{Implementation Details for Baseline Hyperparameter Settings} \label{app:implementation_2}
\paragraph{Flow Matching Model Baselines}
\label{par:flow_implementation}
For all flow matching evaluations, we utilize Stable Diffusion 3.5-Medium as the generative backbone with the time scheduler shift factor set to 4.0 and the number of function evaluations fixed at 50. All experiments are conducted on a single NVIDIA A100 GPU. The specific configurations for each method are as follows:
\begin{itemize}
    \item \textbf{FlowChef} The number of gradient descent steps and the step size serving as the data consistency guidance scale are set to 3 and 1, respectively, via grid search on the calibration set. The CFG scale is fixed at 2.0.
    \item \textbf{FlowDPS} 
    FlowDPS is evaluated in its standard configuration with 50 NFE and a CFG scale of 2, using a step size of 15 with 3 gradient descent steps for the inpainting task and a step size of 10 with 3 gradient descent steps for all remaining tasks.
    \item \textbf{FLAIR} We employ 15 gradient descent steps for data consistency optimization with a CFG scale of 2.0 in its standard configuration. The learning rate is set to 0.1 for deblurring tasks, consistent with the original paper. For SR $\times8$ and SR $\times12$ tasks, the optimal learning rates are determined via grid search on the calibration set: 2 for both tasks on FFHQ, 3 for SR $\times8$ on DIV2K, and 6 for SR $\times12$ on DIV2K.
    \item \textbf{Flower} We employ 15 steps of gradient descent with a step size of 15 for data consistency optimization via a grid search on the calibration set. 
    Consistent with the original authors, the optional uncertainty term is found to destabilize restoration, and thus $\gamma$ is set to 0 with the CFG scale fixed at 2.0.
    \item \textbf{FlowLPS} We employ 6 steps of gradient descent with a step size of 20 for data consistency optimization via a grid search on the calibration set. Consistent with the original paper, we use a fixed number of Langevin-Proximal update steps, $N_p=15$, and a CFG scale of 2.0. A constant step size $\zeta_t = 10^{-4}$ is used for Langevin dynamics across all timesteps. For the proximal optimization, we utilized the SGD optimizer. We use an initial learning rate of 0.5 for super-resolution and 0.1 for other tasks. To ensure stability during optimization, we apply a learning rate decay schedule: for super-resolution, the learning rate is decayed by a factor of 0.85 every 5 iterations, and for inpainting, by 0.65 every 10 iterations.
    \item \textbf{FLINT} {We employ 6 steps of gradient descent for the data consistency update by default, increasing to 9 for DIV2K motion deblurring and to 12 for DIV2K SR$\times$12, with the CFG scale fixed at 1.0. Remaining implementation details follow the setup described in Sec.~\ref{app:implementation_1}.}
\end{itemize}

\paragraph{Diffusion Model Baselines}
\label{par:diffusion_implementation}
For all diffusion-based evaluations, we utilize Stable Diffusion 1.5 as the generative backbone with the NFE fixed at 100. To ensure a competitive comparison, hyperparameters for all baselines are determined via an extensive grid search on a calibration set. All experiments are conducted on a single NVIDIA A100 GPU. Specific configurations for each method are as follows:
\begin{itemize}
    \item \textbf{PSLD} 
    {We use the paper's default loss weights $\omega = 1.0$ (measurement fidelity) and $\gamma = 0.1$ (gluing), applied identically to both super-resolution and motion deblurring. Following the official implementation, the gluing gradient is added to the DDIM update with no additional step size, and $\mathcal{A}^{\!\top}$ approximates the pseudo-inverse for all tasks. We disable classifier-free guidance, following PSLD's official setup with an empty null prompt.}
    \item \textbf{Resample} We adopt the resampling hyperparameter $\gamma \left(\frac{1-\bar\alpha_{t-1}}{\bar\alpha_t}\right)\left(1-\frac{\bar\alpha_t}{\bar\alpha_{t-1}}\right)$ with $\gamma=40$ as proposed in the original paper. To enforce hard data consistency, the skip step size is set to 1, accompanied by 10 steps of gradient descent for SR$\times4$ task and 20 steps of gradient descent for motion deblurring task with a step size of 30. The CFG scale is fixed at 1.0.
    \item \textbf{DDPG} We set hyperparameters for SR$\times4$ and motion deblurring set to $\{\gamma, \zeta, \tilde{\eta}\} = \{10.0, 0.8, 0.3\}$ and $\{5.0, 0.6, 0.6\}$, respectively, with the step size $\mu_t$ set to the theoretically safe value $\mu_t^*$. Here, $\gamma$ controls the guidance transition from back-projection to least-squares, $\zeta$ balances stochastic noise injection against reconstruction accuracy, and $\tilde{\eta}$ regularizes the back-projection operator to prevent noise amplification.
    \item \textbf{TReg} {We follow TReg's algorithmic pipeline (CG-based data-consistency every third sampling step within the first $170$ of $200$ DDIM steps, corresponding to $t \gtrsim 145$ on the standard schedule; adaptive negation of the null prompt; no DPS branch). The numerical hyperparameters $\{\text{CFG scale}, \lambda_{\text{cg}}, \text{lr}_{\text{null}}\} = \{2.0,\ 0.1,\ 10^{-5}\}$ are calibrated for our setting (FFHQ $\times$ SD1.5 $\times$ $\sigma_y = 0.03$, SR$\times 4$ and motion deblurring).}
    \item \textbf{FLINT} {We employ $8$ steps of gradient descent for the data consistency update for super-resolution $\times 4$ and $6$ steps for motion deblurring, with the CFG scale fixed at 1.0. Remaining implementation details follow the setup described in Sec.~\ref{app:implementation_1}.}
\end{itemize}

\section{Additional Experiments} \label{app:add_exp}
\subsection{Additional Task: Inpainting} \label{app:add_exp_1}
\begin{figure*}[htbp!]
  \centering
  \includegraphics[width=0.95\textwidth]{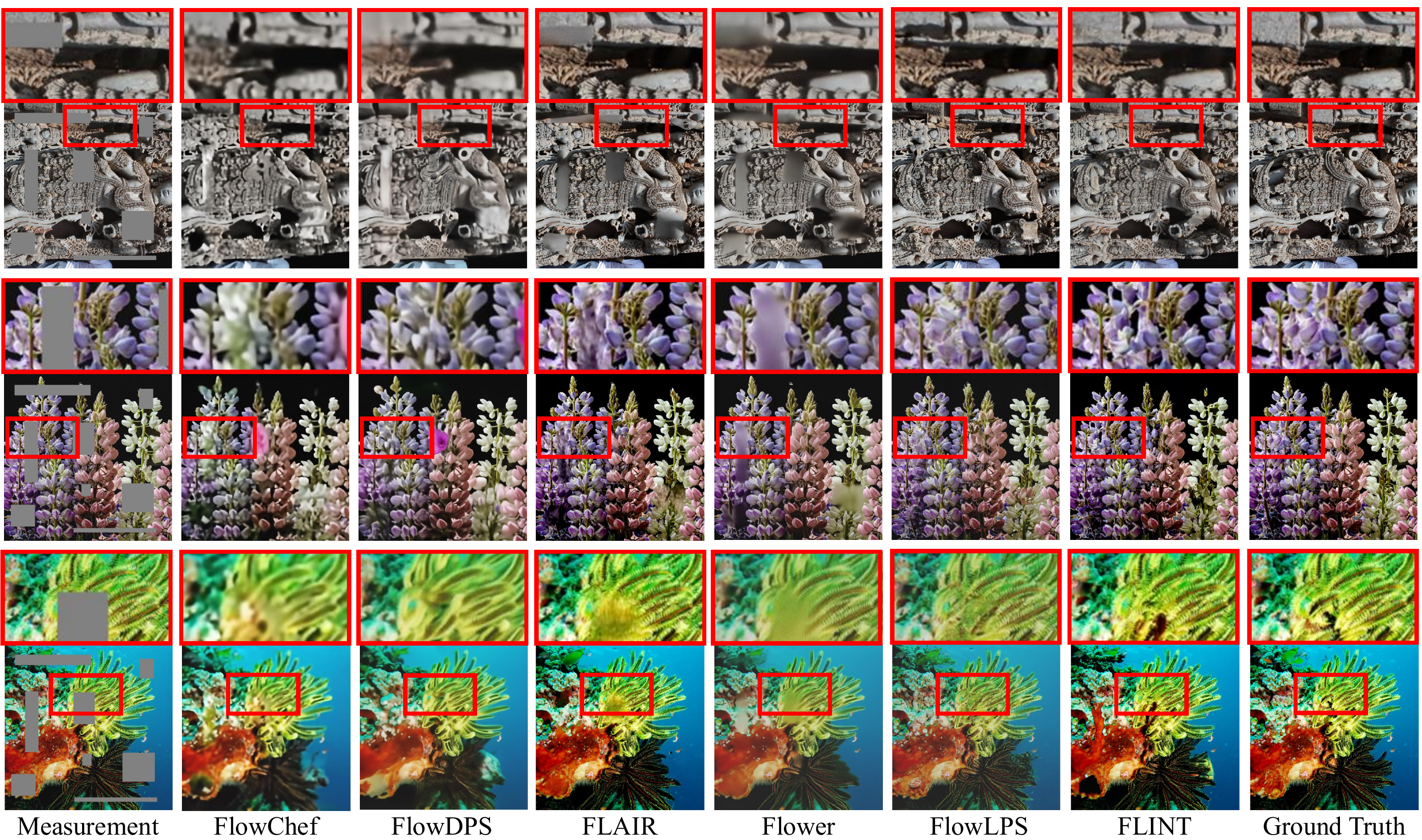}
  \caption{
  Qualitative comparison for DIV2K datasets on inpainting task.
  }
  \label{fig:app_quali_inpainting_flow}
\end{figure*}

Beyond the four inverse problems in Sec.~\ref{par:exp_results_1}, we evaluate FLINT on the inpainting task using a flow matching model. As shown in Table~\ref{tab:quanti_inpainting_flow} and Fig.~\ref{fig:app_quali_inpainting_flow}, FLINT attains state of the art performance on box inpainting across all metrics. Proximal initialization via variational framework combined with guided posterior sampling under a bounded stochasticity schedule suppresses both boundary artifacts and blurry textures prevalent in baseline posterior sampling approaches. 
Moreover, FLINT achieves a favorable position on the perception–distortion trade-off, attaining competitive reconstruction fidelity (PSNR, SSIM) while improving perceptual quality (LPIPS, FID).

\begin{table*}[!ht]
    \centering
    \renewcommand{\arraystretch}{0.9}
    \setlength{\tabcolsep}{0.6em}
    \caption{
    Quantitative comparison on inpainting based on flow matching model. 
    All metrics are averaged over 1,000 FFHQ and 800 DIV2K samples, with each method allocated 50 NFEs under additive Gaussian noise ($\sigma_y = 0.03$).
    }
    \label{tab:quanti_inpainting_flow}
    \resizebox{0.9\textwidth}{!}{
    \begin{tabular}{l|cccc|cccc}
    \toprule
    & \multicolumn{4}{c|}{\textbf{FFHQ (768 $\times$ 768)}}
    & \multicolumn{4}{c}{\textbf{DIV2K (768 $\times$ 768)}}\\
    \midrule
    Method
    & PSNR$\uparrow$ & SSIM$\uparrow$ & FID$\downarrow$ & LPIPS$\downarrow$
    & PSNR$\uparrow$ & SSIM$\uparrow$ & FID$\downarrow$ & LPIPS$\downarrow$ \\
    \midrule
    FlowChef
    & 22.82 & 0.785 & 57.23 & 0.375
    & 20.01 & 0.606 & 70.08 & 0.463 \\
    FlowDPS
    & 21.74 & 0.783 & 33.56 & 0.345
    & 20.36 & 0.618 & 29.97 & 0.415 \\
    FLAIR
    & \textbf{28.34} & \underline{0.867} & 11.86 & \underline{0.141}
    & \textbf{23.74} & \textbf{0.826} & 12.09 & \underline{0.167} \\
    Flower
    & \underline{27.12} & 0.811 & 37.94 & 0.334
    & 22.66 & 0.687 & 36.36 & 0.359 \\
    FlowLPS
    & 26.65 & 0.829 & \textbf{9.59} & 0.249
    & 22.28 & 0.739 & \textbf{8.17} & 0.257 \\
    \midrule
    \rowcolor{red!10}
    Ours
    & 26.52 & \textbf{0.872} & \underline{10.63} & \textbf{0.113}
    & \underline{22.91} & \underline{0.825} & \underline{8.78} & \textbf{0.123} \\
    \bottomrule
    \end{tabular}
    }
\end{table*}

\subsection{Additional Runtime Comparison} \label{app:add_exp_5}
We evaluate the inference efficiency of FLINT by comparing its mean runtime with baselines based on flow matching and diffusion models. As shown in Table~\ref{tab:efficiency_comparison}, we report the per image sampling cost (s/image) averaged over 10 samples.

\begin{table*}[!htbp]
    \centering
    \renewcommand{\arraystretch}{1.0} 
    \setlength{\tabcolsep}{0.4em}    
    \caption{
    Runtime comparison of methods based on flow matching and diffusion models. Sampling time per image, in seconds, is averaged over 10 samples.
    }
    \scriptsize
    \begin{tabular}[t]{l|c|c|c|c}
    \toprule
    \multicolumn{5}{c}{\textbf{Flow Matching Model (SD3.5-M)}} \\ \midrule
    \textbf{Method} 
        & \makecell{SR $\times$8 \\Runtime(s)$\downarrow$} 
        & \makecell{SR $\times$12\\Runtime(s)$\downarrow$} 
        & \makecell{Motion\\Deblur\\Runtime(s)$\downarrow$} 
        & \makecell{Gaussian\\Deblur\\Runtime(s)$\downarrow$} \\ \midrule
    FlowChef    & 8.14 & 8.14 & 9.28 & 8.66 \\
    FlowDPS     & 8.22 & 8.18 & 9.47 & 8.71 \\
    FLAIR       & 20.33 & 20.28 & 22.94 & 21.37 \\
    Flower      & 16.89 & 16.83 & 23.36 & 19.46 \\
    FlowLPS     & 16.84 & 16.78 & 22.82 & 19.31 \\
    Ours (FFHQ) & 6.79 & 6.81 & 6.71 & 7.47 \\
    Ours (DIV2K)& 6.79 & 9.26 & 7.92 & 7.41 \\
    \bottomrule
    \end{tabular}
    \hspace{1.0em}
    \begin{tabular}[t]{l|c|c}
    \toprule
    \multicolumn{3}{c}{\textbf{Diffusion Model (SD1.5)}} \\ \midrule
    \textbf{Method} 
        & \makecell{SR $\times$4\\Runtime(s)$\downarrow$} 
        & \makecell{Motion\\Deblur\\Runtime(s)$\downarrow$} \\ \midrule
    PSLD     & 65.65 & 65.49 \\
    ReSample & 161.29 & 289.24 \\
    DDPG     & 56.98 & 98.04 \\
    TReg     & 18.61 & 18.74 \\
    FLINT \textbf{(Ours)}    & 64.25 & 49.56 \\ \bottomrule
    \end{tabular}
    \label{tab:efficiency_comparison}
\end{table*}

\subsection{Additional Discussion 1: Qualitative Comparison of Sampling Process} \label{app:add_exp_2}
\begin{figure*}[htbp!]
  \centering
  \includegraphics[width=0.95\textwidth]{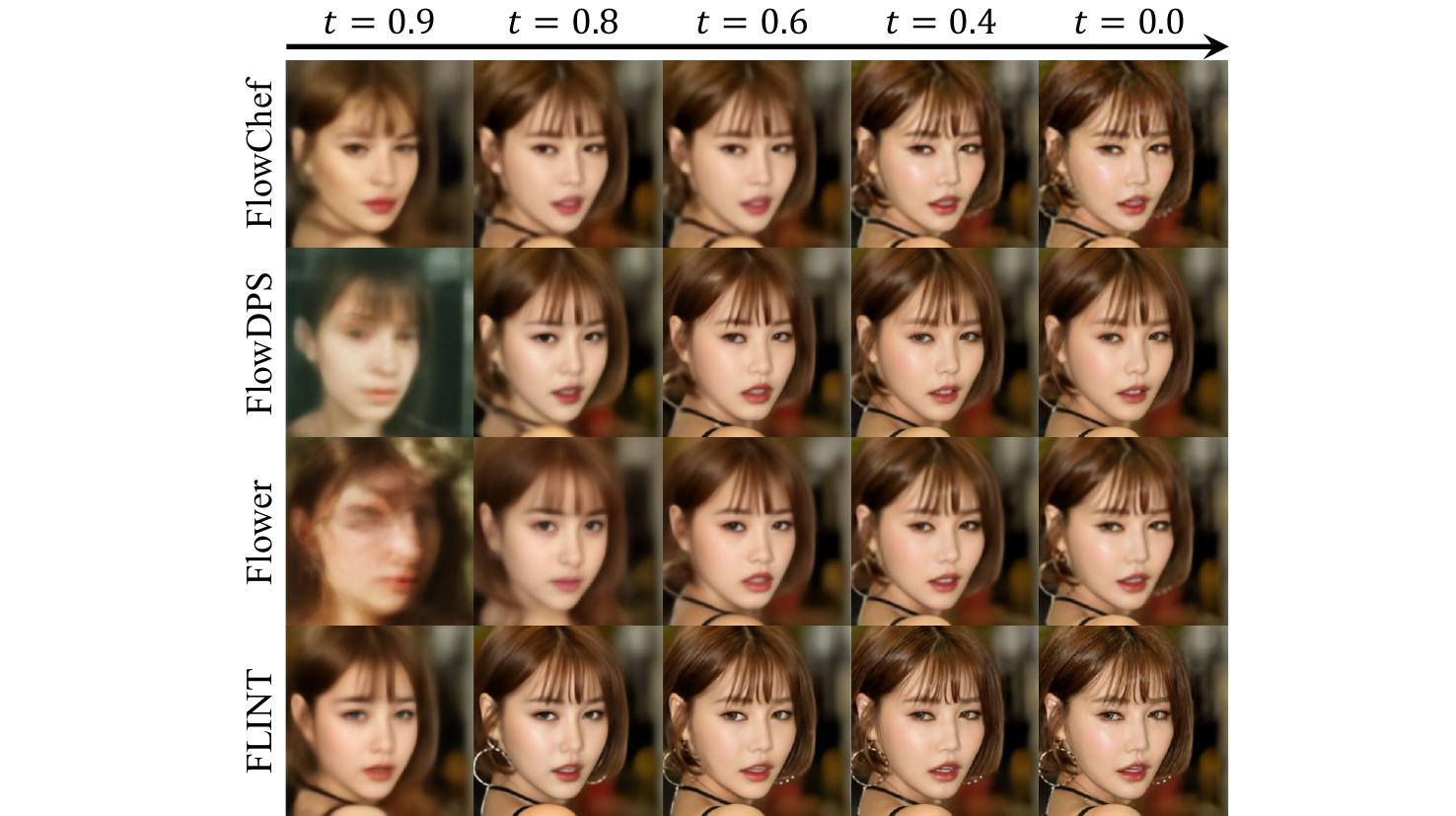}
  \caption{
  Qualitative comparison of the Tweedie estimate's along the sampling trajectory for FLINT and competing baselines.
  }
  \label{fig:app_process_visualization}
\end{figure*}

Figure~\ref{fig:app_process_visualization} visualizes the Tweedie's estimate at representative intermediate timesteps for FLINT and the competing baselines, complementing the quantitative trajectory analysis in Figure~\ref{fig:convergence}. At early timesteps, the baseline Tweedie's estimates remain dominated by noise and exhibit only weak alignment with the underlying measurement, whereas FLINT's estimate already reflects the coarse structure of the target reconstruction. As sampling progresses, FLINT's intermediate estimates progressively refine perceptual detail without disturbing the coarse layout established at initialization, whereas the baselines continue to form the coarse layout itself through much of the sampling trajectory. These patterns visualize, in trajectory form, the two complementary mechanisms identified in Section~\ref{sec:analysis}: proximal initialization places the sampler within a low-energy posterior region from the outset, and the bounded stochasticity schedule preserves this advantage as detail-level refinement proceeds. Together, they allow FLINT to converge along a noticeably shorter path through the generative manifold than the baselines.

\subsection{Additional Discussion 2: Effect of Cumulant Regularization} \label{app:add_exp_7}
The cumulant regularizer $\mathcal{R}(\mu_\phi)$ incorporated into the manifold-constrained proximal objective of stage~1 serves to enforce strict adherence of the optimized noise to the statistical characteristics of the Gaussian prior manifold. While the standard KL divergence primarily constrains lower-order statistics such as mean and variance, this regularization term suppresses non-Gaussian drift by penalizing departures in higher-order moments, including skewness and excess kurtosis. 
Table~\ref{tab:add_exp_cumulant} shows that on motion deblurring (FFHQ, 100 samples), incorporating this term improves perceptual quality metrics, suggesting that controlling higher order statistics keeps the proximal initialization within the valid input support of the pretrained generative model, enabling full utilization of proximity-preserving posterior sampling of stage~2.

\begin{table}[t]
\centering
\caption{Ablation on cumulant regularizer $\mathcal{R}(\mu_\phi)$ coefficient $\lambda_{\mathrm{cum}}$ for Motion Deblurring, averaged over 100 FFHQ samples under Gaussian noise ($\sigma_y = 0.03$).}
\label{tab:ablation_lambda_cumulant_motion}
\begin{tabular}{lcccc}
\toprule
Method & PSNR$\uparrow$ & SSIM$\uparrow$ & KID$\downarrow$ & LPIPS$\downarrow$ \\
\midrule
$\lambda_{\mathrm{cum}}=0$~~(OFF)   & 30.32          & \textbf{0.802} & 0.015          & 0.263 \\
$\lambda_{\mathrm{cum}}=10.0$~(ON)  & \textbf{30.39} & 0.800          & \textbf{0.008} & \textbf{0.224} \\
\bottomrule
\end{tabular}
\label{tab:add_exp_cumulant}
\end{table}

\subsection{Additional Discussion 3: Comparison with Pseudo-Inverse-Based Initialization Strategies} \label{app:add_exp_3}

We compare FLINT against three initialization strategies that derive the starting noise from the pseudo-inverse measurement, with Stage~2 fixed: forward-noising initialization~\cite{meng2022sdedit,chung2022ccdf}, which perturbs the pseudo-inverse via a single linear interpolation; deterministic inversion~\cite{deng2025fireflow}, which integrates the reverse generative ODE; and stochastic inversion~\cite{rout2025semantic}, which integrates a controlled SDE biased toward typical Gaussian noise. None of these approaches jointly satisfies the two proximal conditions: proximity to a measurement-consistent region of source space and consistency with the prior typical set. 

\newpage
\begin{wraptable}{r}{0.51\textwidth}
    \centering
    \renewcommand{\arraystretch}{1.0}
    \setlength{\tabcolsep}{0.6em}
    \vspace{-1em}
    \caption{
    Quantitative comparison of alternative initialization strategies that derive the initial noise from the pseudo-inverse measurement, with Stage~2 fixed. Gaussian deblurring (FFHQ, 100 samples).
    }
    \resizebox{0.5\textwidth}{!}{
        \begin{tabular}{l|cccc}
        \toprule
        & \multicolumn{4}{c}{Gaussian Deblurring} \\
        \midrule
        Method &
        PSNR$\uparrow$ & SSIM$\uparrow$ & KID$\downarrow$ & LPIPS$\downarrow$ \\
        \midrule
        Forward-Noising & 28.64 & 0.616 & 0.020 & 0.449 \\
        Deterministic-Inv. & 29.97 & 0.788 & 0.021 & 0.231 \\
        Stochastic-Inv. & 29.81 & 0.751 & 0.010 & 0.209 \\
        \rowcolor{red!10}
        Ours & \textbf{30.11} & \textbf{0.788} & \textbf{0.005} & \textbf{0.192} \\
        \bottomrule
        \end{tabular}
    }
    \label{tab:add_exp_3}
\end{wraptable}

The forward-noising path does not coincide with the generative trajectory and satisfies neither condition. Deterministic inversion faithfully reproduces the pseudo-inverse but propagates the spectral distortions of the degradation operator into source space, yielding atypical noise. Stochastic inversion enforces typicality through its controller, yet the measurement operator enters neither the control objective nor the dynamics, causing the optimized noise to drift toward a generic typical sample without measurement awareness. FLINT's variational formulation enforces both conditions jointly through a closed-form prior-matching term and a data-fidelity term that explicitly involves the measurement operator. Table~\ref{tab:add_exp_3} confirms this analysis, with FLINT attaining the best score across every metric.

\subsection{Additional Discussion 4: Excessive Source-Space Optimization} \label{app:add_exp_4}
\begin{figure}[!htbp]
    \centering
    \includegraphics[width=\linewidth]{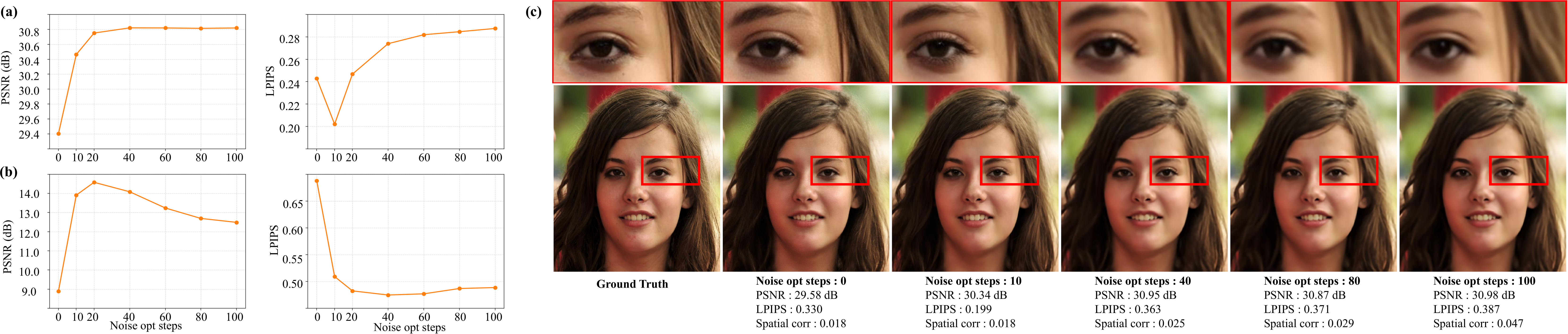}
    \caption{
    Effect of excessive source-space optimization on Gaussian deblurring (FFHQ, 100 samples). (a) PSNR and LPIPS of FLINT reconstructions across noise optimization steps. Beyond the proximal regime (10 steps), PSNR plateaus while LPIPS degrades by over $35\%$, as the OOD noise can no longer be rendered into natural image structures by the pretrained generative model. (b) PSNR and LPIPS of unconditional sampling rendered directly from the optimized noise (NFE\,=\,29, matched to Stage 2). (c) PSNR, LPIPS, and horizontal lag-1 spatial correlation of the optimized noise on a representative sample. The correlation remains stable within the proximal regime and trends upward beyond it, indicating a gradual drift toward the boundary of the prior's typical set.
    }
    \vspace{-1em}
    \label{fig:noiseopt_excessive}
\end{figure}

This section examines the effect of prolonged source-space optimization on reconstruction quality, corroborating the proximal-level optimization budget adopted in Stage 1. Excessive source-space optimization pushes the initial noise out-of-distribution (OOD), reducing reconstruction quality. To quantify this, we track the horizontal lag-1 spatial correlation of the optimized noise across optimization steps. As shown in Figure~\ref{fig:noiseopt_excessive}(c), the correlation remains stable within the proximal regime and gradually increases beyond it, eventually exceeding the $95\%$ acceptance threshold for i.i.d. $\mathcal{N}(0, I)$, indicating that the optimized noise progressively drifts toward the boundary of the prior's typical set. This OOD drift manifests directly in reconstruction quality under guided posterior sampling (Figure~\ref{fig:noiseopt_excessive}(a)). While prolonged optimization yields marginal fidelity gains ($0.36$ dB PSNR from step $10$ to $40$), perceptual quality degrades substantially, with LPIPS increasing by over $35\%$ (from $0.202$ to $0.274$), as the OOD noise can no longer be rendered into natural image structures by the pretrained generative model. These results corroborate the proximal design choice of Stage 1. Prolonged source-space optimization beyond a proximal level yields limited fidelity gains while reducing perceptual quality. A short proximal initialization is therefore preferable to exhaustive source-space optimization.

\subsection{Additional Discussion 5: Posterior Variance} \label{app:add_disc_5}
\begin{figure}[!htbp]
    \centering
    \includegraphics[width=0.9\linewidth]{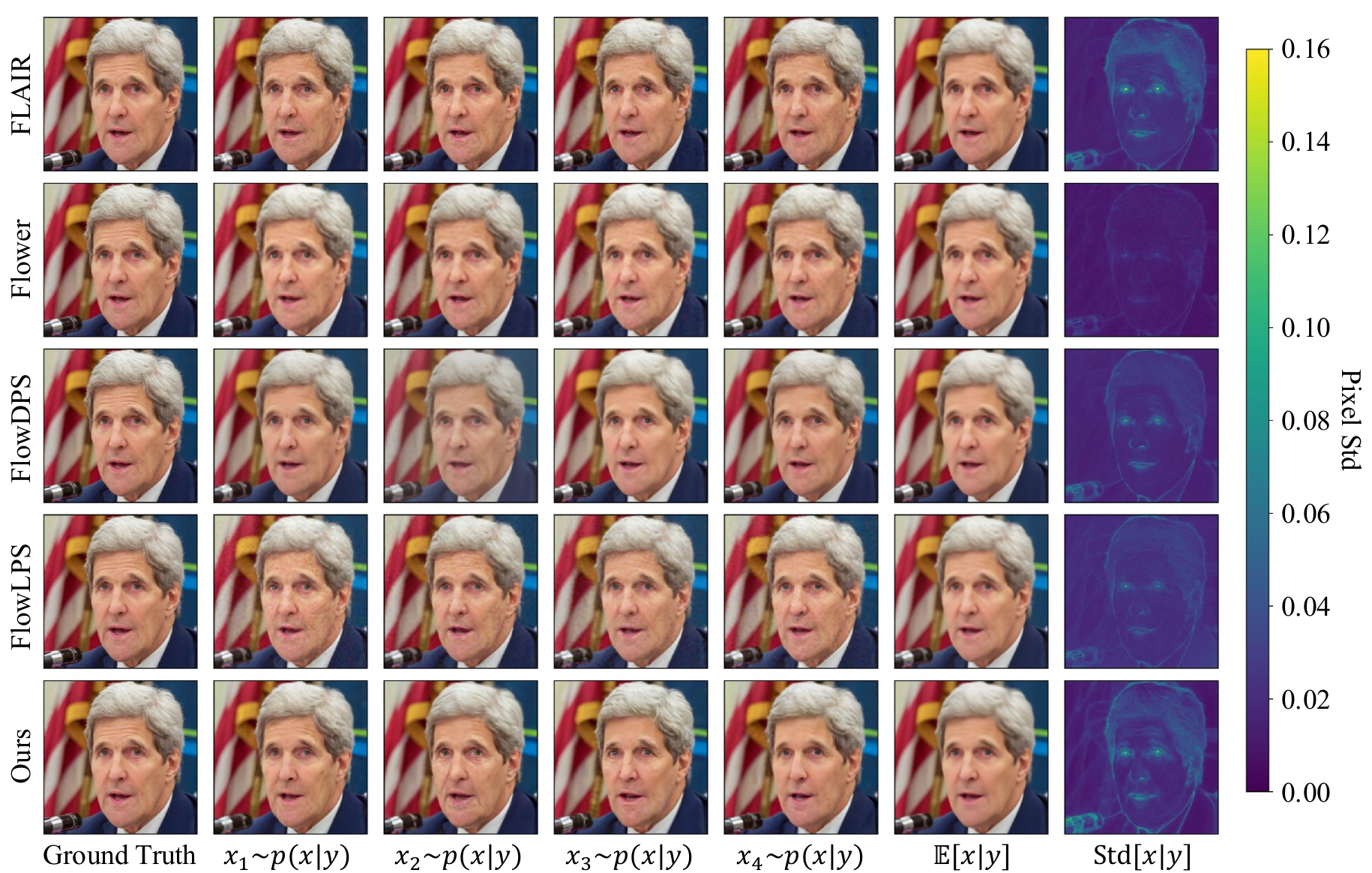}
    \caption{
    Posterior variance analysis for super-resolution $\times$12. For each method, 50 samples are drawn from the posterior $p(\mathbf{x}|\mathbf{y})$ for a fixed observation $\mathbf{y}$, with four representative samples, the posterior mean $\mathbb{E}[\mathbf{x}|\mathbf{y}]$, and the pixel-wise standard deviation $\text{Std}[\mathbf{x}|\mathbf{y}]$ displayed. FLINT produces diverse posterior samples with meaningful variance in perceptually significant regions, confirming the absence of mode collapse.
    }
    \vspace{-1em}
    \label{fig:app_posterior_variance}
\end{figure}

To verify that FLINT does not suffer from mode collapse, we assess the posterior variance $\text{Var}[\mathbf{x}|\mathbf{y}]$ for the super-resolution $\times$12 task by drawing 50 samples for a fixed observation $\mathbf{y}$ and computing their pixel-wise standard deviation. Fig.~\ref{fig:app_posterior_variance} presents this analysis for FLINT alongside FLAIR, Flower, FlowDPS, and FlowLPS. FLINT maintains meaningful posterior coverage with variance concentrated in high-frequency texture regions, precisely the regions left underdetermined by the likelihood term in the highly ill-posed task, confirming that FLINT does not degenerate into a single-mode solution. This diversity arises from two complementary mechanisms. In Stage 1, the spectrally weighted data consistency loss anchors the low-frequency structure of the proximity-initialized noise to the measurement while suppressing gradients from high-frequency components, leaving uncertainty in fine-grained details. Combined with annealed noise in the variational optimization, this encourages the optimizer to explore a range of measurement-consistent proximal points that share the same coarse layout but differ in their high-frequency realization. In Stage 2, the bounded bell-shaped stochasticity schedule concentrates noise injection in intermediate timesteps where perceptual detail is refined, actively generating varying high-frequency components across samples while keeping the low-frequency structure stable. Together, these results confirm that FLINT reliably explores the posterior rather than collapsing to a single reconstruction.

\section{Additional Qualitative Results} \label{app:add_exp_6}
\begin{figure*}[!htbp]
  \centering
  \includegraphics[width=0.79\textwidth]{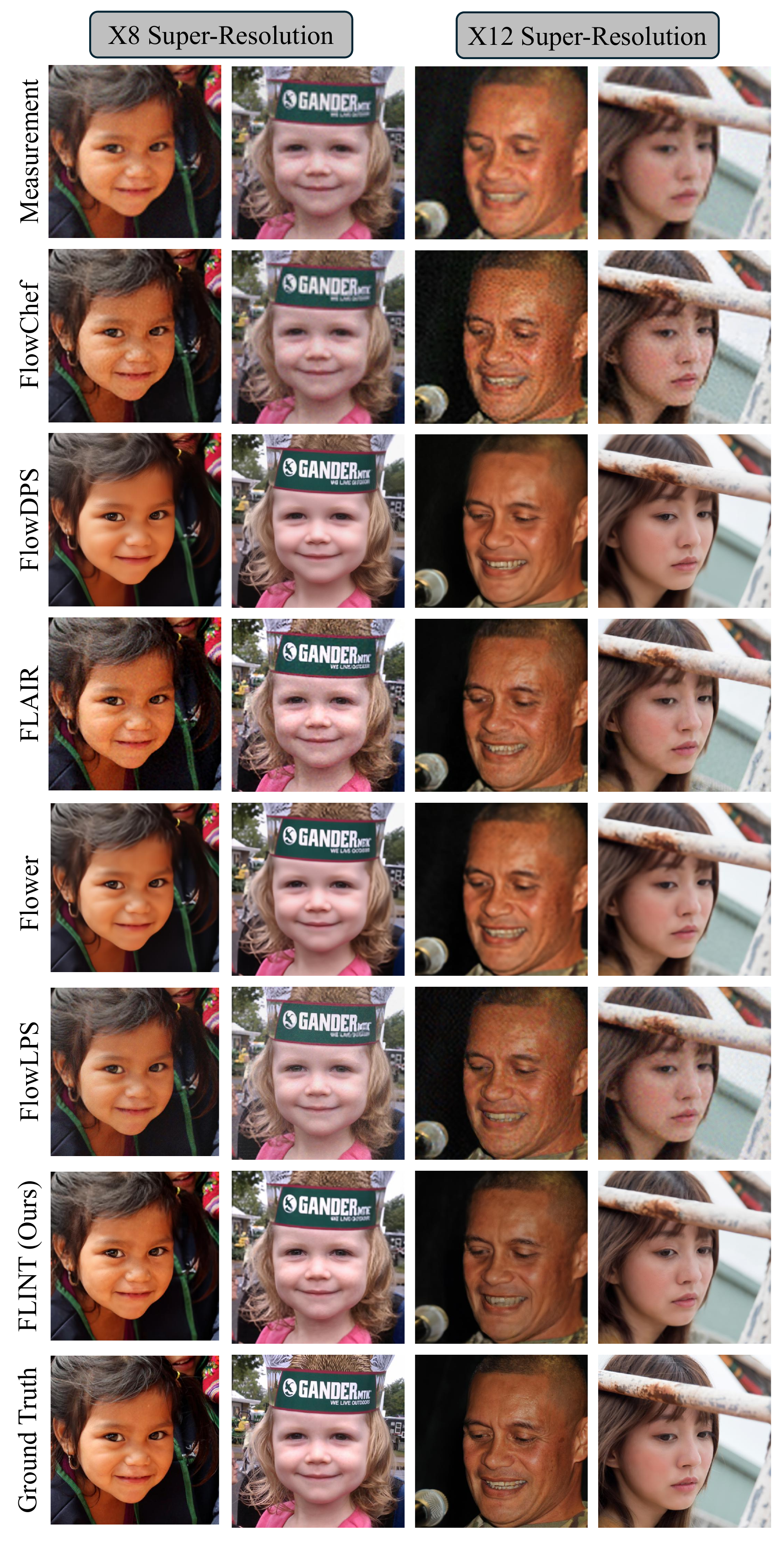}
  \caption{
  Additional qualitative comparison for the super resolution tasks (super-resolution $\times$8 and super-resolution $\times$12) on the FFHQ dataset.
  }
  \label{fig:Add_app_quali_SR_FFHQ}
\end{figure*}

\begin{figure*}[!htbp]
  \centering
  \includegraphics[width=0.79\textwidth]{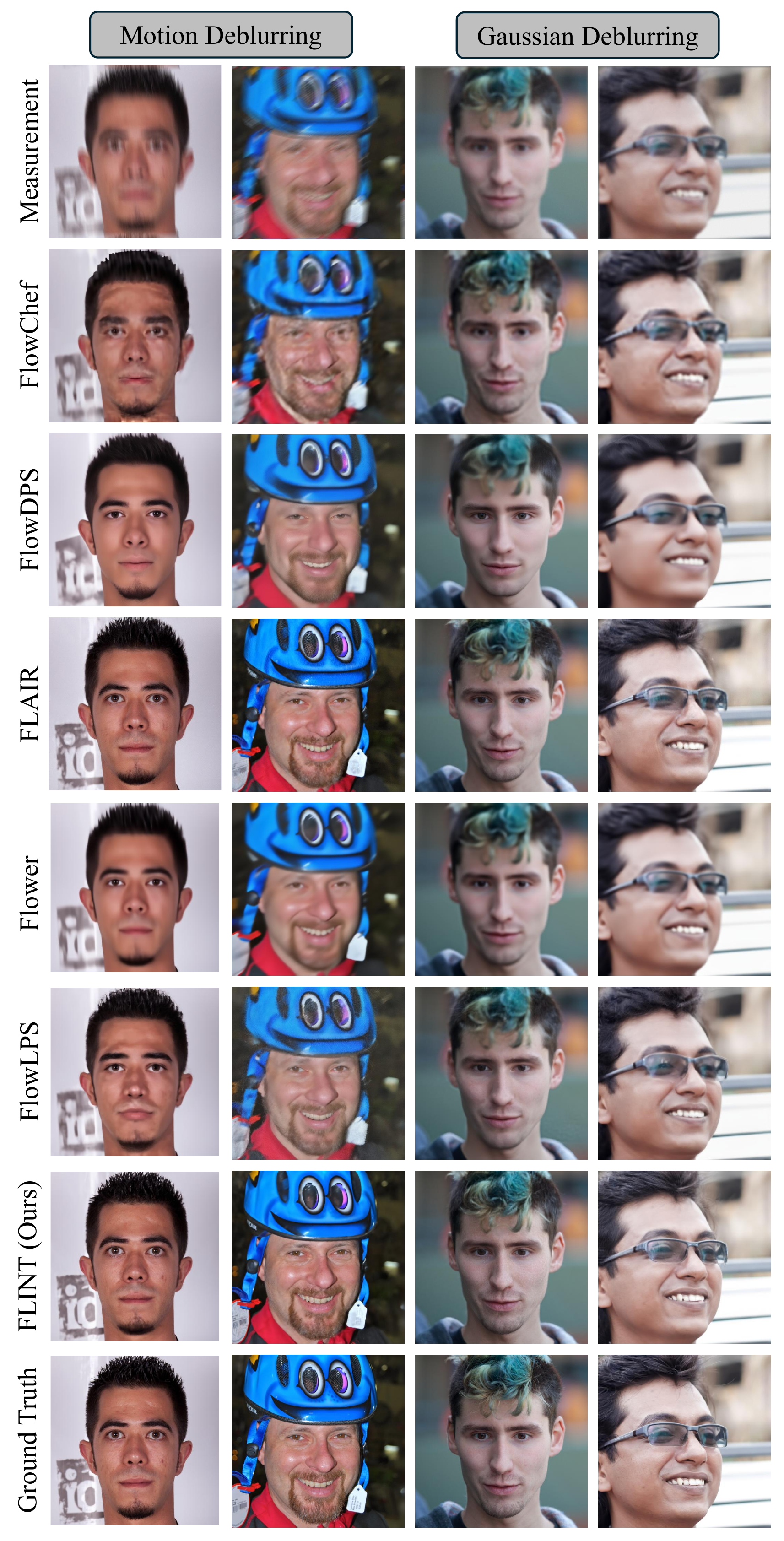}
  \caption{
  Additional qualitative comparison for the deblurring tasks (motion deblurring and gaussian deblurring) on the FFHQ dataset.
  }
  \label{fig:Add_app_quali_Deblur_FFHQ}
\end{figure*}

\begin{figure*}[!htbp]
  \centering
  \includegraphics[width=0.79\textwidth]{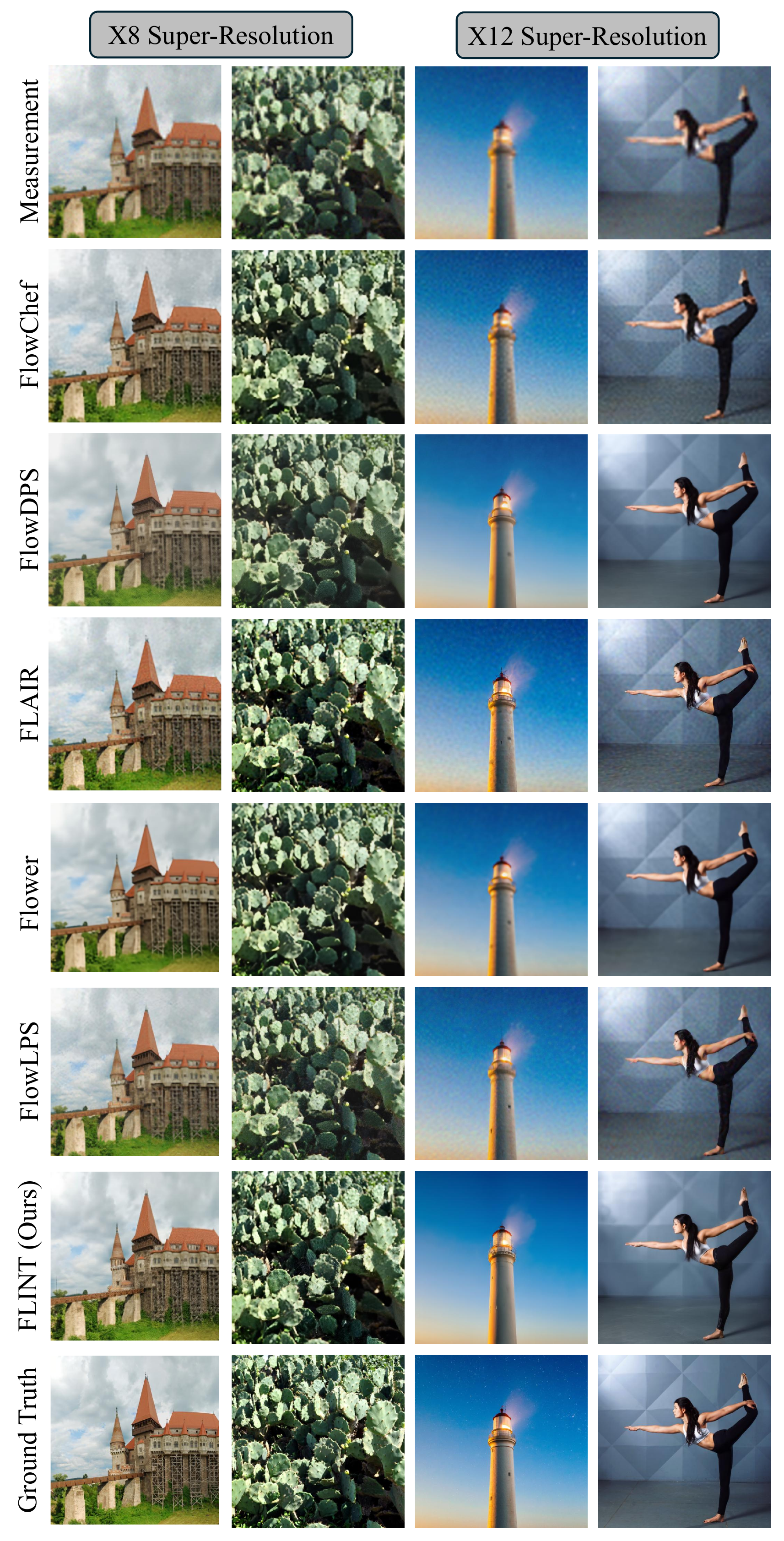}
  \caption{
  Additional qualitative comparison for the super resolution tasks (super-resolution $\times$8 and super-resolution $\times$12) on the DIV2K dataset.
  }
  \label{fig:Add_app_quali_SR_DIV2K}
\end{figure*}

\begin{figure*}[!htbp]
  \centering
  \includegraphics[width=0.79\textwidth]{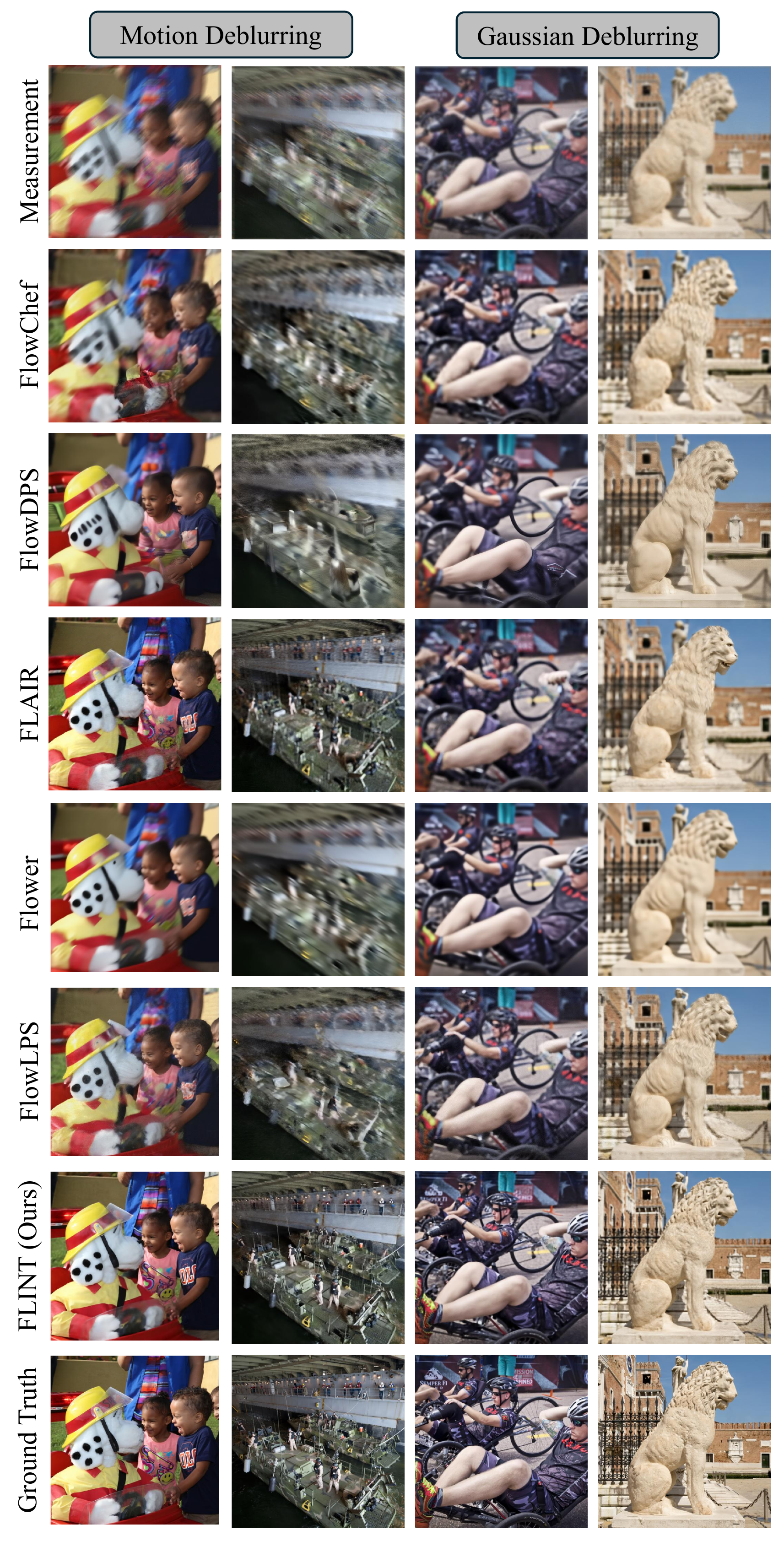}
  \caption{
  Additional qualitative comparison for the deblurring tasks (motion deblurring and gaussian deblurring) on the DIV2K dataset.
  }
  \label{fig:Add_app_quali_Deblur_DIV2K}
\end{figure*}
\newpage

\begin{figure*}[!htbp]
  \centering
  \includegraphics[width=0.81\textwidth]{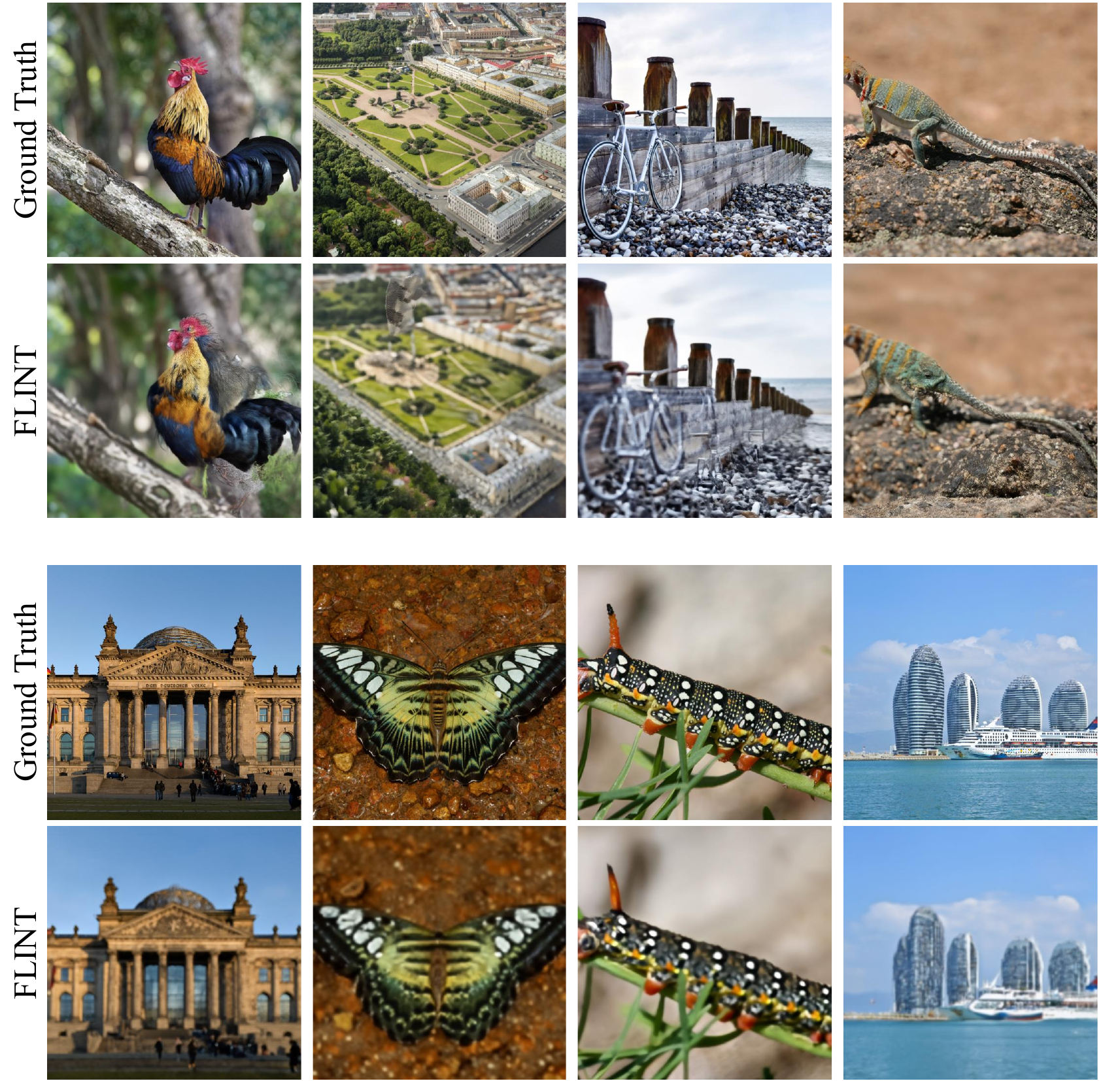}
  \caption{
  Qualitative failure cases of FLINT on highly complex scenes (DIV2K dataset). Increased ill-posedness in dense, multi-object images disrupts the balance between measurement fidelity and the generative prior. This results in either hallucinated structures, where the model generates spurious details to satisfy the prior, or blurry textures, where conflicting updates fail to resolve high-frequency information. These cases highlight the inherited limitations of the pretrained backbone's capacity for complex out-of-distribution scenes.
  }
  \label{fig:appendix_failure_case}
\end{figure*}

\newpage
\section*{NeurIPS Paper Checklist}

\begin{enumerate}

\item {\bf Claims}
    \item[] Question: Do the main claims made in the abstract and introduction accurately reflect the paper's contributions and scope?
    \item[] Answer: \answerYes{} %
    \item[] Justification: The Abstract and Introduction (Sec.~\ref{sec:intro}) clearly reflect the paper’s contributions and scope.

\item {\bf Limitations}
    \item[] Question: Does the paper discuss the limitations of the work performed by the authors?
    \item[] Answer: \answerYes{} %
    \item[] Justification: This paper discusses the limitations in Sec.~\ref{sec:conclusion}.

\item {\bf Theory assumptions and proofs}
    \item[] Question: For each theoretical result, does the paper provide the full set of assumptions and a complete (and correct) proof?
    \item[] Answer: \answerYes{} %
    \item[] Justification: This paper presents a theoretical analysis comprising \textbf{Proposition 1} in Sec.~\ref{sec:analysis} of the main paper, with their corresponding mathematical proofs and assumptions provided in Appendix Subsec.~\ref{app:subsec_energy_advantage}, Subsec.~\ref{app:subsec_proof_error_decomposition}. 

    \item {\bf Experimental result reproducibility}
    \item[] Question: Does the paper fully disclose all the information needed to reproduce the main experimental results of the paper to the extent that it affects the main claims and/or conclusions of the paper (regardless of whether the code and data are provided or not)?
    \item[] Answer: \answerYes{} %
    \item[] Justification: This paper provides sufficient information that can reproduce the results of the main experiment in Sec.~\ref{sec:experiments}, Sec.~\ref{app:implementation}, and additional experiments in Sec.~\ref{app:flint_diffusion}, Sec.~\ref{app:add_exp}.

\item {\bf Open access to data and code}
    \item[] Question: Does the paper provide open access to the data and code, with sufficient instructions to faithfully reproduce the main experimental results, as described in supplemental material?
    \item[] Answer: \answerYes{} %
    \item[] Justification: We will release code and sample data to reproduce the experimental results in Sec.~\ref{sec:experiments}  upon acceptance of the paper. %

\item {\bf Experimental setting/details}
    \item[] Question: Does the paper specify all the training and test details (e.g., data splits, hyperparameters, how they were chosen, type of optimizer) necessary to understand the results?
    \item[] Answer: \answerYes{} %
    \item[] Justification: This paper provides the details of the experiments sufficiently in Sec.~\ref{sec:experiments}, Sec.~\ref{app:implementation}, and Sec.~\ref{app:flint_diffusion}.

\item {\bf Experiment statistical significance}
    \item[] Question: Does the paper report error bars suitably and correctly defined or other appropriate information about the statistical significance of the experiments?
    \item[] Answer: \answerYes{} %
    \item[] Justification: This paper offers several tables for quantitative comparison and figures for qualitative comparison in the main paper as well as the Appendix.

\item {\bf Experiments compute resources}
    \item[] Question: For each experiment, does the paper provide sufficient information on the computer resources (type of compute workers, memory, time of execution) needed to reproduce the experiments?
    \item[] Answer: \answerYes{} %
    \item[] Justification: This paper provides the details of computing resources and time of execution in Fig.~\ref{fig:robustness_nfe} and Tab.~\ref{tab:efficiency_comparison}.
    
\item {\bf Code of ethics}
    \item[] Question: Does the research conducted in the paper conform, in every respect, with the NeurIPS Code of Ethics \url{https://neurips.cc/public/EthicsGuidelines}?
    \item[] Answer: \answerYes{} %
    \item[] Justification: This paper fully complies with the NeurIPS code of ethics.

\item {\bf Broader impacts}
    \item[] Question: Does the paper discuss both potential positive societal impacts and negative societal impacts of the work performed?
    \item[] Answer: \answerYes{} %
    \item[] Justification: In Sec.~\ref{sec:intro} and Sec.~\ref{sec:conclusion} of the main paper, we emphasize that the our work opens a new perspective on the design of flow-based inverse problem solvers, that combines source space warmup initialization via variational framework with guided sampling under a bounded stochasticity schedule.
    
\item {\bf Safeguards}
    \item[] Question: Does the paper describe safeguards that have been put in place for responsible release of data or models that have a high risk for misuse (e.g., pre-trained language models, image generators, or scraped datasets)?
    \item[] Answer: \answerNA{} %
    \item[] Justification: This paper poses no such risks.

\item {\bf Licenses for existing assets}
    \item[] Question: Are the creators or original owners of assets (e.g., code, data, models), used in the paper, properly credited and are the license and terms of use explicitly mentioned and properly respected?
    \item[] Answer: \answerYes{} %
    \item[] Justification: We have fully cited all methods, datasets, and pre-trained foundation models utilized in this paper.

\item {\bf New assets}
    \item[] Question: Are new assets introduced in the paper well documented and is the documentation provided alongside the assets?
    \item[] Answer: \answerYes{} %
    \item[] Justification: The assets introduced in this paper have been documented, and we publicly plan to release the code in the future.

\item {\bf Crowdsourcing and research with human subjects}
    \item[] Question: For crowdsourcing experiments and research with human subjects, does the paper include the full text of instructions given to participants and screenshots, if applicable, as well as details about compensation (if any)? 
    \item[] Answer: \answerNA{} %
    \item[] Justification: Our study does not involve crowdsourcing or research with human subjects.

\item {\bf Institutional review board (IRB) approvals or equivalent for research with human subjects}
    \item[] Question: Does the paper describe potential risks incurred by study participants, whether such risks were disclosed to the subjects, and whether Institutional Review Board (IRB) approvals (or an equivalent approval/review based on the requirements of your country or institution) were obtained?
    \item[] Answer: \answerNA{} %
    \item[] Justification: No new data collection or direct interaction with human subjects was conducted in this study.

\item {\bf Declaration of LLM usage}
    \item[] Question: Does the paper describe the usage of LLMs if it is an important, original, or non-standard component of the core methods in this research? Note that if the LLM is used only for writing, editing, or formatting purposes and does \emph{not} impact the core methodology, scientific rigor, or originality of the research, declaration is not required.
    \item[] Answer: \answerNA{} %
    \item[] Justification: The core method on this paper does not involve LLMs as any important, original, or non-standard components.

\end{enumerate}

\end{document}